\pdfoutput=1

\documentclass[11pt]{article}

\usepackage[final]{acl}
\usepackage{kotex}
\usepackage{amsmath}
\usepackage{times}
\usepackage[table]{xcolor}
\usepackage{latexsym}
\usepackage{booktabs}
\usepackage{multirow}
\usepackage{booktabs}
\usepackage{multirow}
\usepackage{amssymb}
\usepackage{pifont}
\usepackage{makecell}
\newcommand{\xmark}{\ding{55}}
\usepackage[T1]{fontenc}

\usepackage[utf8]{inputenc}

\usepackage{microtype}

\usepackage{inconsolata}

\usepackage{graphicx}

\title{RL-ARC: Calibrating Large Reasoning Models via Reasoning-guided Uncertainty}

\author{
Gukhyeon Lee$^{1}$ \quad
SangKeun Lee$^{1, 2}$ \\
$^{1}$Department of Artificial Intelligence, Korea University, Seoul, Republic of Korea \\
$^{2}$Department of Computer Science and Engineering, Korea University, Seoul, Republic of Korea \\
\texttt{\{gukhyeon\_lee, yalphy\}@korea.ac.kr}
}

\begin{document}
\maketitle
\begin{abstract}
Language models (LMs) are commonly trained with Reinforcement Learning with Verifiable Rewards (RLVR) to enhance their reasoning capabilities. However, since RLVR does not explicitly account for calibration during training, it can lead to severe calibration degradation, including overconfidence. Recent calibration-aware training methods for LMs, which incorporate objectives for uncertainty estimation into training, improve calibration but still exhibit overconfidence under distribution shift, while sacrificing reasoning performance. To this end, we propose \textsc{RL-ARC}, a calibration-aware training framework that jointly leverages reasoning confidence and answer confidence. Specifically, \textsc{RL-ARC} leverages reasoning confidence as an auxiliary signal for calibrating answer confidence, applying it as reasoning-guided regularization for correct cases and as an overconfidence penalty for incorrect cases. Comprehensive results across ID and OOD settings show that, beyond improving calibration, \textsc{RL-ARC} enables reasoning models to adaptively estimate confidence based on the given question without substantially sacrificing reasoning performance, thereby highlighting the importance of reasoning confidence for training reliable reasoning models\footnote{Our code and data are available at \url{https://github.com/2gukhyeon/RL-ARC}.}.
\end{abstract}

\section{Introduction}
\begin{figure}[t]
    \centering
    \includegraphics[width=1\linewidth]{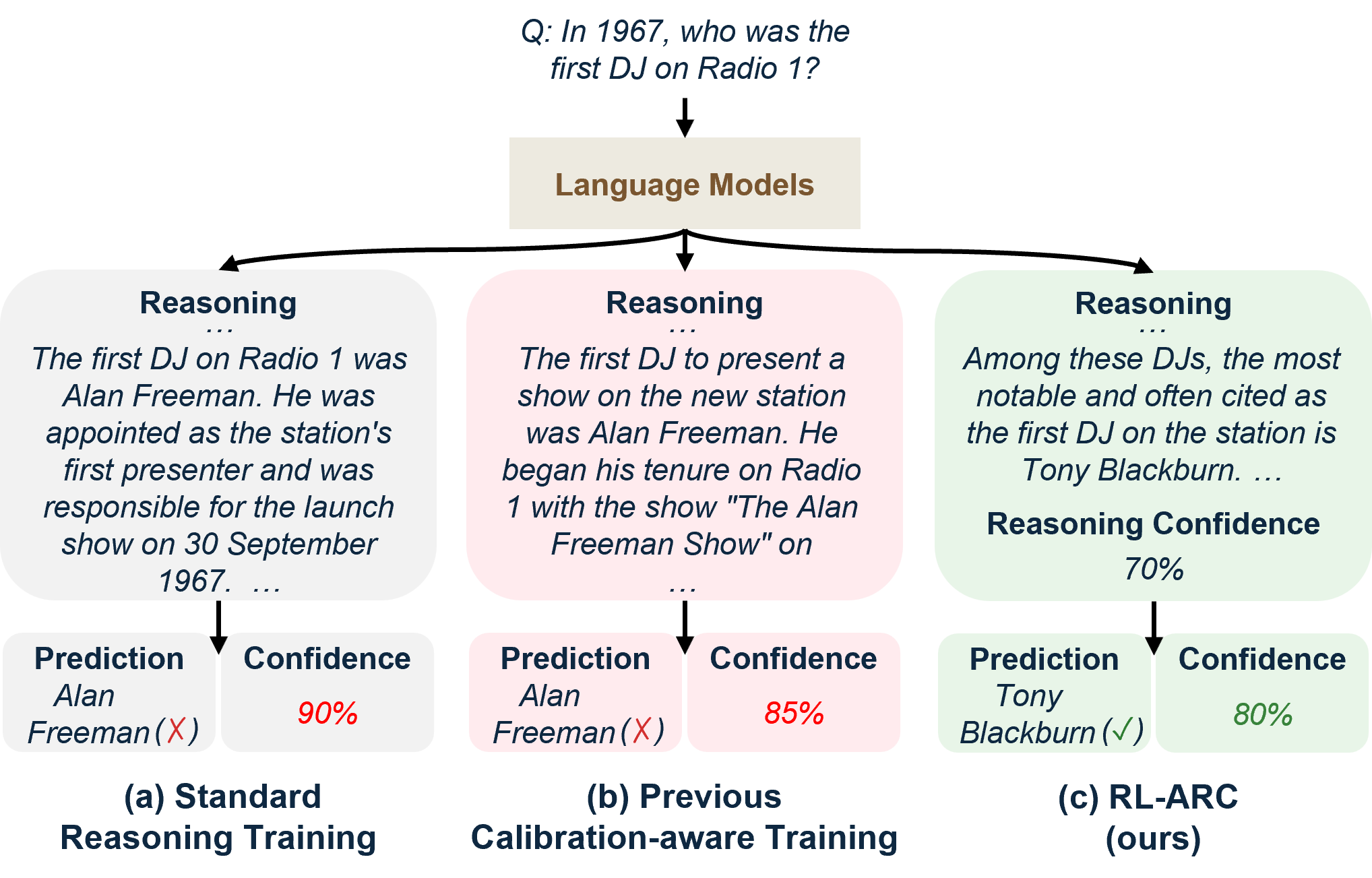}
    \caption{Comparison of reasoning training frameworks: (a) Standard reasoning training often assigns high confidence to incorrect answers (i.e., overconfidence). (b) Previous calibration-aware training reduces overconfidence to some extent, but still tends to assign high confidence to incorrect answers on OOD benchmarks. (c) Our \textsc{RL-ARC} consistently improves calibration while preserving accuracy gains under OOD settings.}
    \label{fig:case_study}
\end{figure}
Large Reasoning Models (LRMs) have achieved remarkable performance on challenging tasks, such as arithmetic reasoning, largely due to their enhanced reasoning capabilities \cite{lewkowycz2022solving, lambert2025tulu, openai2026openaio1card}. These capabilities are primarily attributed to Reinforcement Learning (RL), which enables models to engage in a slow thinking process expressed in natural language \cite{kahneman2011thinking, yang2024qwen25mathtechnicalreportmathematical}. A representative training paradigm for improving reasoning ability is Reinforcement Learning with Verifiable Rewards (RLVR), which provides a simple yet effective framework by relying solely on the correctness of the model’s prediction~\cite{lambert2025tulu}. However, since RL training such as RLVR does not take into account the confidence in its outputs, it tends to reinforce overconfident reasoning \cite{kirichenko2025abstentionbench, mei-etal-2026-reasoning}. This critical limitation restricts the applicability of language models in high-stakes domains, such as healthcare and finance, where the reliability of predictions is essential \cite{lin-etal-2022-truthfulqa}.

One promising direction for addressing this limitation is to explicitly reward models for estimating their uncertainty \cite{leng2025taming, liu2025c2gspgconfidencecalibratedgroupsequence, damani2026beyond}. By optimizing an objective function that jointly considers the correctness of the model’s prediction and the calibration of its uncertainty, these approaches can effectively improve the reliability of model outputs. However, these approaches focus primarily on calibrating answer confidence, without considering the reliability of the reasoning process itself.
As a result, reasoning models remain vulnerable to generating unreliable or incorrect reasoning, regardless of whether their predictions are correct or reliable~\cite{khalid-etal-2025-large, chen2026can}. 
This limitation can further lead to overconfidence issues, where models assign high confidence to the predictions supported by flawed reasoning or even to incorrect predictions. 
Moreover, directly optimizing calibration objectives often degrades reasoning performance, leading to an accuracy-calibration trade-off~\cite{damani2026beyond}. 
As illustrated in Figure~\ref{fig:case_study}, these issues become more pronounced under distribution shift, where confidence-aware behavior differs between in-distribution (ID) and out-of-distribution (OOD) settings.

To address these limitations, we propose Reinforcement Learning with Auxiliary Reasoning Confidence (\textsc{RL-ARC}), a simple calibration-aware RL framework that jointly leverages reasoning confidence and answer confidence. Specifically, \textsc{RL-ARC} extracts two types of confidence for a given question: reasoning confidence, which reflects the correctness (or truthfulness) of the model’s reasoning process, and answer confidence, which reflects the correctness of the model’s prediction. We then train the model to minimize the discrepancy between its correctness and answer confidence, while modulating the degree of optimization using reasoning confidence.

To validate the effectiveness of \textsc{RL-ARC} in terms of both performance and calibration, we evaluate it on mathematical reasoning benchmarks with varying levels of difficulty, and further extend our evaluation to diverse OOD benchmarks, including complex reasoning and factual QA tasks. Experimental results demonstrate that \textsc{RL-ARC} consistently outperforms existing calibration-aware training methods in calibration, while substantially mitigating accuracy degradation.

The contributions of this work are summarized as follows:
\begin{itemize}
\item[$\circ$] We introduce \textsc{RL-ARC}, a simple yet effective calibration-aware RL framework that jointly leverages reasoning and answer confidence.
\item[$\circ$] We leverage reasoning confidence as an auxiliary signal to regularize answer confidence for correct predictions and penalize overconfidence in incorrect predictions.
\item[$\circ$] We demonstrate that \textsc{RL-ARC} improves calibration over existing reasoning training methods without substantially sacrificing accuracy, thereby highlighting the importance of reasoning confidence in calibration-aware training.
\end{itemize}

\section{Related Work}
\subsection{Uncertainty Estimation for Language Models}
Estimating the uncertainty of language models plays an important role in improving the reliability of their generated responses \cite{xia-etal-2025-survey, yoon2026reasoning}. Existing studies on uncertainty estimation for language models can be broadly categorized into three classes. First, accuracy-based approaches estimate uncertainty using the accuracy of multiple model-generated samples \cite{lin2024generating, xue-etal-2025-ualign}. However, these approaches incur substantial inference costs, and their uncertainty estimates can be unstable depending on the prompt designs used to generate diverse samples \cite{pmlr-v139-zhao21c, tonolini-etal-2024-bayesian}. Another line of work estimates uncertainty from the probability distributions over tokens generated by the language model, known as probability-based approaches \cite{kadavath2022languagemodelsmostlyknow, kuhn2023semantic}. However, these approaches have limited applicability in open-ended generation settings, and mitigating this limitation often requires additional inference costs \cite{kuhn2023semantic}. To address these limitations, recent studies have proposed verbalized confidence-based approaches, where the model explicitly expresses its confidence in its own answer \cite{tian-etal-2023-just, xiong2024can, dong-etal-2024-llm}. These approaches are simple, human-interpretable \cite{yoon2026reasoning}, and correlated with model performance and response quality \cite{tian-etal-2023-just, dong-etal-2024-llm}. However, they still suffer from overconfidence and remain limited in recognizing their own mistakes.

\begin{figure*}[t]
    \centering
    \includegraphics[width=1\linewidth]{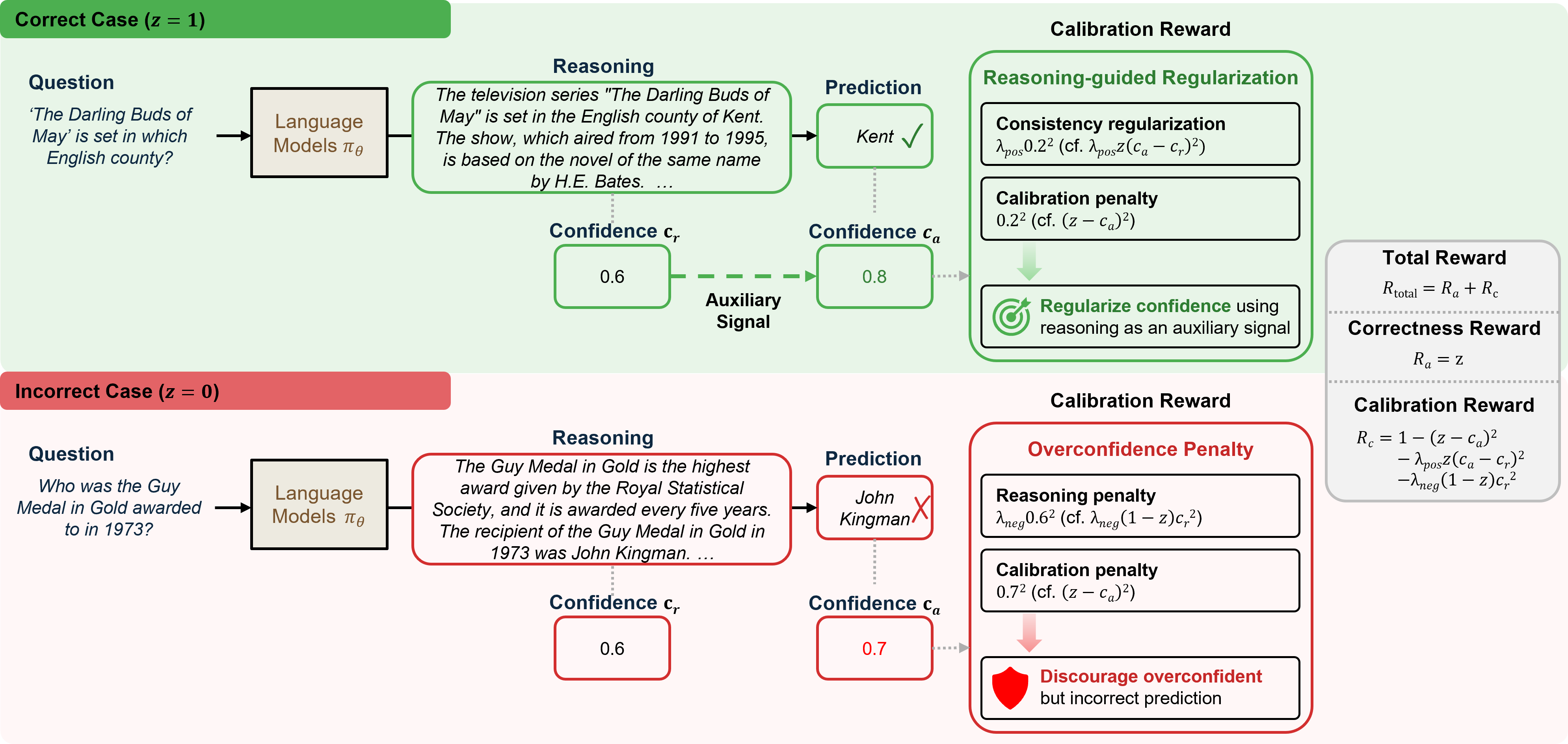}
    \caption{Overview of \textsc{RL-ARC}. \textsc{RL-ARC} leverages reasoning confidence differently depending on whether the prediction is correct. For correct cases, \textsc{RL-ARC} uses reasoning confidence to regularize answer confidence, encouraging high confidence only when both the prediction and the reasoning process are correct. For incorrect cases, \textsc{RL-ARC} penalizes overconfidence by leveraging confidence from both the reasoning process and the prediction.}
    \label{fig:concept_figure}
\end{figure*}

\subsection{Training for Calibration}
Although current LMs have been shown to possess the ability to self-assess their reliability \cite{yoon2026reasoning}, their confidence are often poorly calibrated with the actual correctness of their outputs, suffering from issues such as overconfidence \cite{tian-etal-2023-just, xiong2024can, mei-etal-2026-reasoning} and unstable uncertainty estimation \cite{lin2024generating, xue-etal-2025-ualign}. To address these limitations, several studies have explored calibration-aware training frameworks that incorporate uncertainty estimation objectives into reasoning training \cite{liu2025c2gspgconfidencecalibratedgroupsequence, damani2026beyond, wu2026mitigatingllmhallucinationbehaviorally}. These approaches train LMs to align their confidence with the correctness of their predictions, enabling better-calibrated confidence estimates. However, they still exhibit an accuracy-calibration trade-off and remain prone to overconfidence, particularly in OOD settings.

In contrast, \textsc{RL-ARC} accounts for not only the correctness of the predictions but also the correctness (or truthfulness) of the reasoning processes when aligning self-assessed confidence. This encourages LMs to consider mistakes recognized during reasoning when estimating confidence, going beyond calibration based solely on the correctness of the predictions. As a result, \textsc{RL-ARC} improves calibration across both ID and OOD settings while maintaining competitive accuracy, thereby mitigating the accuracy-calibration trade-off.

\section{RL-ARC: Reinforcement Learning with Auxiliary Reasoning Confidence}
We introduce Reinforcement Learning with Auxiliary Reasoning Confidence (\textsc{RL-ARC}), a calibration-aware RL framework that leverages verbalized confidence from both the reasoning process and the prediction. We first revisit the standard reasoning training algorithm and the proper scoring rule adopted in our method (\S\ref{subsec:preliminaries}), and then describe how we train LMs to improve calibration without substantially sacrificing accuracy (\S\ref{subsec:reasoning-aware_calibration}).

\subsection{Preliminaries}
\label{subsec:preliminaries}
\paragraph{Reinforcement Learning with Verifiable Rewards (RLVR).}
When training LMs $\pi_{\theta}$ to improve their reasoning capabilities, a standard choice of reward function is a binary reward based on prediction correctness:
\begin{equation}
R_{\mathrm{a}}(y, y^*) =
\mathbf{1}_{y=y^*}.
\label{RLVR_reward}
\end{equation}
Here, $y$ is the prediction generated by the LMs $\pi_{\theta}$, $y^*$ is the ground-truth answer, and $\mathbf{1}_{y=y^*}$ is the indicator function that evaluates whether $y$ is correct to $y^*$. RLVR effectively improves the model’s reasoning capability by optimizing LMs $\pi_{\theta}$ to maximize the reward for answer correctness ($R_a$), defined as follows:
\begin{equation}
\max_{\theta} \; \mathbb{E}_{(x,y^*) \sim \mathcal{D},\; y \sim \pi_{\theta}(\cdot \mid x)} \left[ R_a(y, y^*) \right].
\end{equation}

\paragraph{Brier Score.}
Accuracy and calibration represent distinct concepts, yet they can be unified through proper scoring rules such as the Brier Score \cite{Brier1950VERIFICATIONOF, emde2025towards}.
Specifically, Brier Score \cite{Brier1950VERIFICATIONOF} is a widely used proper scoring rule, mathematically defined as the mean squared error between a continuous confidence score and one-hot encoded correctness data. The score is formulated as follows:
\begin{equation}
\mathrm{Brier}(y, c_a, y^{*})
=
\left(\mathbf{1}_{y=y^*} - c_a\right)^{2}.
\end{equation}
Here, $c_a$ denotes the LM $\pi_{\theta}$'s self-assessed confidence in $y$.
By combining it with the RLVR reward function (Eq. equation~\ref{RLVR_reward}), an optimal score can be achieved only when the model predicts both accurately and with appropriate confidence \cite{damani2026beyond}. The objective is defined as follows:
\begin{equation}
\max_{\theta}\;
\mathbb{E}_{(x,y^*) \sim \mathcal{D},\; (y,c_a) \sim \pi_{\theta}(\cdot \mid x)}
\left[ R_{\mathrm{total}}(y, c_a, y^*) \right],
\end{equation}
where \(R_{\mathrm{total}}(y, c_a, y^*) = R_a - \mathrm{Brier}(y, c_a, y^*)\).

\subsection{Reasoning-aware Calibration}
\label{subsec:reasoning-aware_calibration}
The main goal of \textsc{RL-ARC} is to train LMs to generate consistently calibrated outputs not only in ID settings but also in OOD settings, while mitigating the accuracy-calibration trade-off.
\paragraph{Confidence Elicitation.}
To optimize LMs for generating calibrated outputs, \textsc{RL-ARC} first prompts the LM to estimate verbalized confidence in both its reasoning process and prediction for the given question. In this work, we define reasoning confidence as the self-assessed confidence that the reasoning process is correct (or truthful) for solving the given question, and answer confidence as self-assessed confidence that its prediction is correct. These confidence are generated alongside the model’s reasoning and prediction using a calibration-aware reasoning training prompt that instructs the LMs to explicitly estimate confidence while solving the given question (see Appendix~\ref{appendix:Calibration-aware Reasoning Training Prompt Template} for more details about the prompts).

\paragraph{Calibration-aware Optimization.}
By leveraging explicitly elicited confidence in both the reasoning process and the model’s prediction, \textsc{RL-ARC} trains LMs through calibration-aware reinforcement learning to improve calibration while preserving reasoning performance. This leads to the following calibration-aware objective:
\begin{equation}
\label{our total objecive function}
\max_{\theta} \;
\mathbb{E}_{(x,y^*) \sim \mathcal{D},\; (y,c_a,c_r) \sim \pi_{\theta}(\cdot \mid x)}
\left[
R_{\mathrm{total}}(z, c_a, c_r)
\right],
\end{equation}
where
\[
\begin{aligned}
R_{\mathrm{total}}(z,c_a,c_r)
&= R(z,c_a)-\Omega_{\mathrm{arc}}(z,c_a,c_r), \\
R(z,c_a)
&= z-(z-c_a)^2, \\
z
&= \mathbf{1}_{y=y^*}.
\end{aligned}
\]
Here, $c_r$ is the model’s self-assessed confidence associated with its reasoning process. By optimizing the model to maximize the reward for reasoning-aware calibration (Eq. equation~\ref{our total objecive function}), \textsc{RL-ARC} effectively improves calibration while mitigating accuracy-calibration trade-off\footnote{In Section~\ref{experiments}, we demonstrate that applying different training objectives based on prediction correctness is crucial for mitigating the accuracy–calibration trade-off.}. Specifically, \textsc{RL-ARC} leverages reasoning confidence to calibrate LMs with an objective conditioned on whether the model’s prediction is correct, as defined below:
\begin{equation}
\Omega_{\mathrm{arc}}(z, c_a, c_r)
=
\begin{cases}
\lambda_{\mathrm{pos}}(c_a-c_r)^2, & z=1\\
\lambda_{\mathrm{neg}}c_r^2, & z=0.
\end{cases}
\end{equation}
Here, \(\lambda_{\mathrm{pos}}\) and \(\lambda_{\mathrm{neg}}\) are hyperparameters controlling the strength of reasoning confidence $c_r$ to the calibration objective. As shown in Figure~\ref{fig:concept_figure}, when the prediction is correct, \textsc{RL-ARC} encourages the model to assign high answer confidence $c_a$ to its prediction, while also minimizing its misalignment with reasoning confidence $c_r$. This prevents the model from assigning high answer confidence solely based on the prediction, and instead encourages high answer confidence only when both the prediction and the underlying reasoning process are correct. Conversely, for incorrect predictions, \textsc{RL-ARC} penalizes the model for assigning high answer confidence $c_a$ to its prediction. The reasoning confidence $c_r$ in the reasoning process that led to the incorrect prediction is further used as an additional penalty, encouraging the model to calibrate its answer confidence $c_a$ by accounting for the correctness of both the reasoning process and the prediction. Consequently, \textsc{RL-ARC} discourages high answer confidence $c_a$ in incorrect outputs, thereby mitigating overconfidence. Intuitively, the training objective of \textsc{RL-ARC} incentivizes correctness while penalizing the model for assigning high answer confidence $c_a$ to incorrect outputs or low answer confidence $c_a$ to correct outputs, taking into account both the prediction and the reasoning process (Eq. equation~\ref{our total objecive function}).

\section{Experiments}
\label{experiments}
We evaluate \textsc{RL-ARC} to verify its effectiveness in both performance and reliability. Specifically, we aim to answer the following research questions:
\begin{itemize}
\item[$\circ$] Does \textsc{RL-ARC} improve calibration while mitigating the accuracy-calibration trade-off across both ID and OOD settings?
\item[$\circ$] Can \textsc{RL-ARC} adaptively estimate confidence according to the given question?
\item[$\circ$] Does \textsc{RL-ARC} encourage LMs to generate more reliable reasoning?
\end{itemize}

\begin{table*}[ht]
\centering
\resizebox{\textwidth}{!}{
\begin{tabular}{llcccccccc}
\toprule
\multirow{2}{*}{Distribution} & \multirow{2}{*}{Method}
& \multicolumn{4}{c}{Qwen2.5 (7B)}
& \multicolumn{4}{c}{Qwen3 (8B)} \\
\cmidrule(lr){3-6} \cmidrule(lr){7-10}
& & Acc.\,($\uparrow$) & AUROC\,($\uparrow$) & Brier\,($\downarrow$) & \cellcolor{teal!10}ECE\,($\downarrow$)
& Acc.\,($\uparrow$) & AUROC\,($\uparrow$) & Brier\,($\downarrow$) & \cellcolor{teal!10}ECE\,($\downarrow$) \\
\midrule

\multirow{7}{*}{ID (Avg.)}
& Base & 41.19 & 0.58 & 0.54 & \cellcolor{teal!10}0.54 & 55.63 & 0.51 & 0.42 & \cellcolor{teal!10}0.42 \\
& \multicolumn{1}{l}{RLVR} \\        
& \hspace{0.4em}w/ Confidence & \textbf{56.86} & 0.45 & 0.41 & \cellcolor{teal!10}0.40 & \textbf{66.58} & 0.66 & 0.31 & \cellcolor{teal!10}0.32 \\
&\hspace{0.4em}w/ Probability & \textbf{56.86}  & \underline{0.60}  & 0.41 & \cellcolor{teal!10}0.41 & \textbf{66.58}  & 0.76  & 0.27 & \cellcolor{teal!10}0.29 \\
&\hspace{0.4em}w/ Post-hoc & \textbf{56.86}  & \textbf{0.72}  & \underline{0.24} & \cellcolor{teal!10}\underline{0.21} & \textbf{66.58}  & \textbf{0.83}  & \underline{0.16} & \cellcolor{teal!10}0.20 \\
& Behavioral Calibration & 54.75 & 0.50 & 0.27 & \cellcolor{teal!10}0.23 & 65.96 & 0.71 & 0.18 & \cellcolor{teal!10}0.17 \\
& RLCR & 54.82 & 0.58 & 0.25 & \cellcolor{teal!10}0.22 & 65.98 & \underline{0.76} & 0.17 & \cellcolor{teal!10}\underline{0.16} \\
& \cellcolor[gray]{.9}\textsc{RL-ARC} (ours) & \cellcolor[gray]{.9}\underline{56.53} & \cellcolor[gray]{.9}\underline{0.60} & \cellcolor[gray]{.9}\textbf{0.23} & \cellcolor{teal!10}\textbf{0.18} & \cellcolor[gray]{.9}\underline{66.10} & \cellcolor[gray]{.9}\textbf{0.83} & \cellcolor[gray]{.9}\textbf{0.13} & \cellcolor{teal!10}\textbf{0.15} \\

\midrule

\multirow{7}{*}{OOD (Avg.)}
& Base & 43.52 & \underline{0.53} & 0.49 & \cellcolor{teal!10}0.49 & \underline{51.37} & 0.65 & 0.34 & \cellcolor{teal!10}0.34 \\
& \multicolumn{1}{l}{RLVR} \\        
& \hspace{0.4em}w/ Confidence & \textbf{47.58} & 0.50 & 0.50 & \cellcolor{teal!10}0.50 & 47.24 & 0.55 & 0.47 & \cellcolor{teal!10}0.48 \\
&\hspace{0.4em}w/ Probability & \textbf{47.58} & 0.51 & 0.48 & \cellcolor{teal!10}0.47 & 47.24  & 0.50 & 0.45 & \cellcolor{teal!10}0.45 \\
&\hspace{0.4em}w/ Post-hoc & \textbf{47.58}  & 0.51  & 0.31 & \cellcolor{teal!10}0.24  & 47.24  & 0.57  & 0.28  & \cellcolor{teal!10}0.24  \\
& Behavioral Calibration & 47.04 & 0.51 & 0.30 & \cellcolor{teal!10}0.22 & 51.35 & 0.60 & 0.33 & \cellcolor{teal!10}0.31 \\
& RLCR & 46.02 & \textbf{0.54} & \underline{0.28} & \cellcolor{teal!10}\underline{0.20} & 50.99 & \underline{0.66} & \underline{0.26} & \cellcolor{teal!10}\underline{0.22} \\
& \cellcolor[gray]{.9}\textsc{RL-ARC} (ours) & \cellcolor[gray]{.9}\underline{47.27} & \cellcolor[gray]{.9}\textbf{0.54} & \cellcolor[gray]{.9}\textbf{0.26} & \cellcolor{teal!10}\textbf{0.16} & \cellcolor[gray]{.9}\textbf{51.61} & \cellcolor[gray]{.9}\textbf{0.69} & \cellcolor[gray]{.9}\textbf{0.23} & \cellcolor{teal!10}\textbf{0.17} \\

\bottomrule
\end{tabular}
}
\caption{Evaluation results for accuracy (\%) and calibration metrics on Math (ID) and OOD benchmarks. The best and second-best results are highlighted in \textbf{boldface} and \underline{underlined}.}
\label{tabel: main results}
\end{table*}

\subsection{Experimental Setups}
\paragraph{Datasets.}
Following the setup of previous work \cite{damani2026beyond}, all LMs in our main experiments are trained on the Big-Math \cite{albalak2025bigmathlargescalehighqualitymath} dataset. We evaluate \textsc{RL-ARC} on various arithmetic reasoning benchmarks that are similar to the training distribution, including Big-Math \cite{albalak2025bigmathlargescalehighqualitymath}, GSM8K \cite{cobbe2021trainingverifierssolvemath}, MATH-500 \cite{hendrycks2021measuring}, AMC23\footnote{https://huggingface.co/datasets/math-ai/amc23}, AIME24\footnote{https://huggingface.co/datasets/math-ai/aime24}, and AIME25\footnote{https://huggingface.co/datasets/math-ai/aime25}. We further evaluate \textsc{RL-ARC} on benchmarks from distributions different from the training environment, including complex reasoning benchmarks such as StrategyQA \cite{geva-etal-2021-aristotle}, GPQA \cite{rein2024gpqa}, and HotpotQA \cite{yang-etal-2018-hotpotqa}, as well as representative factual QA benchmarks such as SimpleQA \cite{wei2024measuringshortformfactualitylarge}, NQ-Open \cite{kwiatkowski-etal-2019-natural}, and TriviaQA \cite{joshi-etal-2017-triviaqa}. To further validate generalization across training distributions, we additionally use the HotpotQA \cite{yang-etal-2018-hotpotqa} dataset as an alternative training dataset and evaluate the trained models on arithmetic reasoning and factual QA benchmarks. Further details on these datasets are provided in Appendix~\ref{appendix;dataset_details}.
\paragraph{Baselines.}

We compare \textsc{RL-ARC} with various confidence estimation and calibration methods, including Base, RLVR with Confidence, RLVR with Probability, RLVR with Post-hoc, RLCR, and Behavioral Calibration. Base extracts verbalized confidence from the pre-trained model. RLVR with Confidence, such as \cite{xiong2024can}, extracts verbalized confidence from a model trained with standard RLVR. RLVR with Probability, such as \cite{kuhn2023semantic}, computes the average probability of prediction tokens from a model trained with standard RLVR. RLVR with Post-hoc, such as \cite{li-etal-2025-towards}, extracts confidence using an external confidence estimator trained to predict the confidence of the target model’s responses. Behavioral Calibration \cite{wu2026mitigatingllmhallucinationbehaviorally} is a calibration-aware RL approach that utilizes an uncertainty obtained by aggregating verbalized confidence expressed throughout the reasoning process.
RLCR \cite{damani2026beyond} is a representative calibration-aware RL approach that leverages verbalized confidence associated with model predictions.

\begin{table*}[ht]
\centering
\small
\setlength{\tabcolsep}{4pt}
\resizebox{\textwidth}{!}{
\begin{tabular}{llcccccccc}
\toprule
\multirow{2}{*}{Backbone}
& \multirow{2}{*}{Method}
& \multicolumn{4}{c}{ID (Avg.)}
& \multicolumn{4}{c}{OOD (Avg.)} \\
\cmidrule(lr){3-6} \cmidrule(lr){7-10}
& 
& Acc.\,($\uparrow$) 
& AUROC\,($\uparrow$) 
& Brier\,($\downarrow$) 
& \cellcolor{teal!10}ECE\,($\downarrow$)
& Acc.\,($\uparrow$) 
& AUROC\,($\uparrow$) 
& Brier\,($\downarrow$) 
& \cellcolor{teal!10}ECE\,($\downarrow$) \\
\midrule

\multirow{4}{*}{Qwen2.5-7B}
& Base
& 41.19 & \underline{0.58} & 0.54 & \cellcolor{teal!10}0.54
& 43.52 & \underline{0.53} & 0.49 & \cellcolor{teal!10}0.49 \\

& RLVR
& \textbf{56.86} & 0.45 & 0.41 & \cellcolor{teal!10}0.40
& \textbf{47.58} & 0.50 & 0.50 & \cellcolor{teal!10}0.50 \\

& RLCR
& 54.82 & \underline{0.58} & \underline{0.25} & \cellcolor{teal!10}\underline{0.22}
& 46.02 & \textbf{0.54} & \underline{0.28} & \cellcolor{teal!10}\underline{0.20} \\

& \cellcolor[gray]{.9}\textsc{RL-ARC}
& \cellcolor[gray]{.9}\underline{56.53} & \cellcolor[gray]{.9}\textbf{0.60} & \cellcolor[gray]{.9}\textbf{0.23} & \cellcolor{teal!10}\textbf{0.18}
& \cellcolor[gray]{.9}\underline{47.27} & \cellcolor[gray]{.9}\textbf{0.54} & \cellcolor[gray]{.9}\textbf{0.26} & \cellcolor{teal!10}\textbf{0.16} \\

\midrule

\multirow{4}{*}{\makecell[l]{Llama-8B\\(DeepSeek-R1 Distilled)}}
& Base
& 33.24 & 0.61 & 0.62 & \cellcolor{teal!10}0.63
& \textbf{45.96} & \underline{0.61} & 0.43 & \cellcolor{teal!10}0.44 \\

& RLVR
& \textbf{59.49} & 0.51 & 0.36 & \cellcolor{teal!10}0.35
& 40.48 & 0.52 & 0.53 & \cellcolor{teal!10}0.53 \\

& RLCR
& 52.06 & \underline{0.80} & \underline{0.29} & \cellcolor{teal!10}\underline{0.29}
& 42.14 & \textbf{0.64} & \underline{0.30} & \cellcolor{teal!10}\underline{0.28} \\

& \cellcolor[gray]{.9}\textsc{RL-ARC}
& \cellcolor[gray]{.9}\underline{56.69} & \cellcolor[gray]{.9}\textbf{0.81} & \cellcolor[gray]{.9}\textbf{0.27} & \cellcolor{teal!10}\textbf{0.25}
& \cellcolor[gray]{.9}\underline{43.30} & \cellcolor[gray]{.9}\textbf{0.64} & \cellcolor[gray]{.9}\textbf{0.25} & \cellcolor{teal!10}\textbf{0.19} \\

\bottomrule
\end{tabular}
}
\caption{
Evaluation results across different backbone models on Math (ID) and OOD benchmarks.
The best and second-best results within each model are highlighted in
\textbf{boldface} and \underline{underlined}, respectively.
}
\label{tab:backbone_generalization}
\end{table*}

\paragraph{Training Details.}
We utilize the Qwen2.5-7B pre-trained model \cite{qwen2025qwen25technicalreport} to evaluate the effectiveness of \textsc{RL-ARC} as a reasoning training framework for improving both accuracy and calibration. We further utilize the Qwen3-8B model \cite{yang2025qwen3technicalreport}, a recent state-of-the-art language model from the Qwen family, to assess its effectiveness on a stronger backbone. To showcase the generalizability of \textsc{RL-ARC} across model families, we additionally conduct experiments on a Llama-8B distilled from DeepSeek-R1 \cite{Guo_2025}. We employ GRPO \cite{shao2024deepseekmathpushinglimitsmathematical} as the base reinforcement learning algorithm. Further training details are provided in Appendix~\ref{appendix:implmentation}.

\paragraph{Evaluation Details.}
To demonstrate the effectiveness of \textsc{RL-ARC} in improving both reasoning performance and reliability, we evaluate it using various metrics for measuring performance and calibration, including Accuracy, Brier Score (Brier), Expected Calibration Error (ECE), Area Under the ROC Curve (AUROC), and Area Under the Risk-Coverage Curve (AURC). We also evaluate the reliability of the reasoning content through Reasoning Calibration and Shallow Reasoning (SR) metrics. Further evaluation details are provided in Appendix~\ref{appendix:metrics_details}.

\begin{table*}[ht]
\centering
\small
\setlength{\tabcolsep}{4pt}
\resizebox{\textwidth}{!}{
\begin{tabular}{llcccccccc}
\toprule
\multirow{2}{*}{Training Distribution}
& \multirow{2}{*}{Method}
& \multicolumn{4}{c}{ID (Avg.)}
& \multicolumn{4}{c}{OOD (Avg.)} \\
\cmidrule(lr){3-6} \cmidrule(lr){7-10}
&
& Acc.\,($\uparrow$)
& AUROC\,($\uparrow$)
& Brier\,($\downarrow$)
& \cellcolor{teal!10}ECE\,($\downarrow$)
& Acc.\,($\uparrow$)
& AUROC\,($\uparrow$)
& Brier\,($\downarrow$)
& \cellcolor{teal!10}ECE\,($\downarrow$) \\
\midrule

\multirow{4}{*}{Arithmetic Reasoning}
& Base
& 41.19 & \underline{0.58} & 0.54 & \cellcolor{teal!10}0.54
& 43.52 & \underline{0.53} & 0.49 & \cellcolor{teal!10}0.49 \\

& RLVR
& \textbf{56.86} & 0.45 & 0.41 & \cellcolor{teal!10}0.40
& \textbf{47.58} & 0.50 & 0.50 & \cellcolor{teal!10}0.50 \\

& RLCR
& 54.82 & \underline{0.58} & \underline{0.25} & \cellcolor{teal!10}\underline{0.22}
& 46.02 & \textbf{0.54} & \underline{0.28} & \cellcolor{teal!10}\underline{0.20} \\

& \cellcolor[gray]{.9}\textsc{RL-ARC}
& \cellcolor[gray]{.9}\underline{56.53} & \cellcolor[gray]{.9}\textbf{0.60} & \cellcolor[gray]{.9}\textbf{0.23} & \cellcolor{teal!10}\textbf{0.18}
& \cellcolor[gray]{.9}\underline{47.27} & \cellcolor[gray]{.9}\textbf{0.54} & \cellcolor[gray]{.9}\textbf{0.26} & \cellcolor{teal!10}\textbf{0.16} \\

\midrule

\multirow{4}{*}{Complex Reasoning}
& Base
& 39.10 & 0.54 & 0.53 & \cellcolor{teal!10}0.54
& \underline{45.41} & \underline{0.53} & 0.48 & \cellcolor{teal!10}0.49 \\

& RLVR
& \textbf{56.30} & 0.50 & 0.43 & \cellcolor{teal!10}0.43
& 39.51 & 0.50 & 0.60 & \cellcolor{teal!10}0.60 \\

& RLCR
& 54.70 & \underline{0.57} & \underline{0.25} & \cellcolor{teal!10}\underline{0.07}
& 43.62 & \textbf{0.57} & \underline{0.27} & \cellcolor{teal!10}\underline{0.20} \\

& \cellcolor[gray]{.9}\textsc{RL-ARC}
& \cellcolor[gray]{.9}\underline{55.10} & \cellcolor[gray]{.9}\textbf{0.59} & \cellcolor[gray]{.9}\textbf{0.24} & \cellcolor{teal!10}\textbf{0.06}
& \cellcolor[gray]{.9}\textbf{45.42} & \cellcolor[gray]{.9}\textbf{0.57} & \cellcolor[gray]{.9}\textbf{0.25} & \cellcolor{teal!10}\textbf{0.16} \\

\bottomrule
\end{tabular}
}
\caption{
Evaluation results across different training distributions.
The best and second-best results within each training distribution are highlighted in
\textbf{boldface} and \underline{underlined}, respectively.
}
\label{tab:distribution_generalization}
\end{table*}

\begin{table*}[ht]
\centering
\small
\setlength{\tabcolsep}{4pt}
\resizebox{\textwidth}{!}{
\begin{tabular}{lcccccccc|cccc}
\toprule
\multirow{2}{*}{Method}

& \multicolumn{4}{c}{Complex Reasoning}
& \multicolumn{4}{c|}{Factual QA}
& \multicolumn{4}{c}{Avg.} \\

\cmidrule(lr){2-5}
\cmidrule(lr){6-9}
\cmidrule(lr){10-13}

& Acc.\,($\uparrow$)
& AUROC\,($\uparrow$)
& Brier\,($\downarrow$)
& \cellcolor{teal!10}ECE\,($\downarrow$)

& Acc.\,($\uparrow$)
& AUROC\,($\uparrow$)
& Brier\,($\downarrow$)
& \cellcolor{teal!10}ECE\,($\downarrow$)

& Acc.\,($\uparrow$)
& AUROC\,($\uparrow$)
& Brier\,($\downarrow$)
& \cellcolor{teal!10}ECE\,($\downarrow$) \\

\midrule

Base
& 47.95 & \textbf{0.55} & 0.44 & \cellcolor{teal!10}0.43
& 39.10 & \underline{0.51} & 0.55 & \cellcolor{teal!10}0.56
& 43.52 & \underline{0.53} & 0.49 & \cellcolor{teal!10}0.49 \\

RLVR
& \textbf{54.86} & 0.50 & 0.45 & \cellcolor{teal!10}0.45
& \underline{40.30} & 0.50 & 0.56 & \cellcolor{teal!10}0.56
& \textbf{47.58} & 0.50 & 0.50 & \cellcolor{teal!10}0.50 \\

RLCR
& 51.83 & 0.53 & \underline{0.27} & \cellcolor{teal!10}\underline{0.16}
& 40.22 & \textbf{0.54} & \underline{0.29} & \cellcolor{teal!10}\underline{0.23}
& 46.02 & \textbf{0.54} & \underline{0.28} & \cellcolor{teal!10}\underline{0.20} \\

\cellcolor[gray]{.9}\textsc{RL-ARC}
& \cellcolor[gray]{.9}\underline{53.36} & \cellcolor[gray]{.9}\underline{0.54} & \cellcolor[gray]{.9}\textbf{0.25} & \cellcolor{teal!10}\textbf{0.14}
& \cellcolor[gray]{.9}\textbf{41.19} & \cellcolor[gray]{.9}\textbf{0.54} & \cellcolor[gray]{.9}\textbf{0.27} & \cellcolor{teal!10}\textbf{0.19}
& \cellcolor[gray]{.9}\underline{47.27} & \cellcolor[gray]{.9}\textbf{0.54} & \cellcolor[gray]{.9}\textbf{0.26} & \cellcolor{teal!10}\textbf{0.16} \\
\bottomrule
\end{tabular}
}
\caption{
Evaluation results for accuracy (\%) and calibration metrics on six OOD benchmarks, including three complex reasoning and three factual question answering benchmarks. The best and second-best results are highlighted in \textbf{boldface} and \underline{underlined}. Here, we use Qwen2.5-7B as base model.
}
\label{tab:generalization_results}
\end{table*}

\begin{figure*}[ht]
    \centering
    \includegraphics[width=1\linewidth]{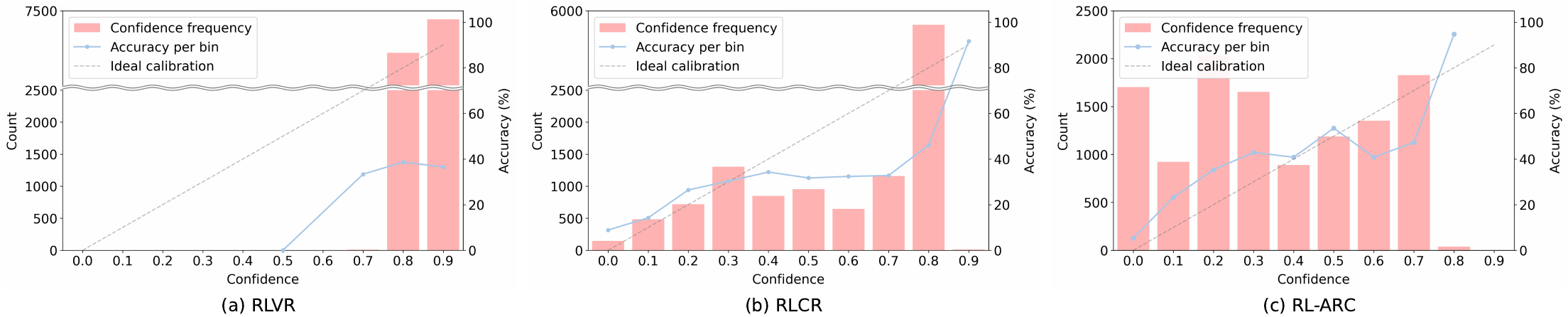}
    \caption{Performance (solid lines) and confidence frequency (bars) across confidence bins on OOD benchmarks for Qwen2.5-7B trained with each method.}
    \label{fig:main_ece_graph}
\end{figure*}

\subsection{Main Results}
To validate the efficacy of \textsc{RL-ARC} as a calibration-aware training method, we first compare it with various uncertainty estimation and calibration baselines. As shown in Table~\ref{tabel: main results}, \textsc{RL-ARC} consistently outperforms these methods in calibration capability, while achieving competitive performance comparable to that of RLVR method, which is trained solely to improve accuracy, across both ID and OOD benchmarks. Notably, \textsc{RL-ARC} consistently outperforms these methods in calibration by a substantial margin on OOD benchmarks, without substantially sacrificing reasoning performance across both evaluated models. Although the existing post-hoc calibration method achieves competitive calibration on ID benchmarks by additionally training an estimator on the ID dataset, we observe that its effectiveness does not consistently transfer to OOD benchmarks. This suggests that an external confidence estimator trained on the ID dataset may have limited generalization under distribution shifts. The existing calibration-aware training methods, on the other hand, generalize robustly to OOD benchmarks in terms of calibration, but often incur noticeable degradation in reasoning performance. In contrast, \textsc{RL-ARC} robustly and effectively improves calibration while largely preserving the performance gains from training, indicating that incorporating reasoning confidence into calibration-aware training provides an effective way to mitigate the accuracy-calibration trade-off. More detailed results are provided in Appendix~\ref{appendix:overall_results_for_each_task}.

\subsection{General Applicability to Architectures and Training Distributions}
\paragraph{Applicability across Model Families.}
To validate the broader applicability of \textsc{RL-ARC}, we evaluate its effectiveness across diverse model families, including Qwen- and Llama-based models. As shown in Table~\ref{tab:backbone_generalization}, \textsc{RL-ARC} consistently improves calibration while substantially mitigating accuracy degradation across both model families. Notably, although Llama-8B distilled from DeepSeek-R1 still exhibits accuracy degradation on OOD benchmarks when trained with \textsc{RL-ARC}, similar to other training methods, \textsc{RL-ARC} is effective at mitigating the degradation in accuracy. The results demonstrate that \textsc{RL-ARC}, which leverages confidence in the reasoning process, not only effectively improves calibration but also substantially mitigates accuracy degradation across diverse model families.
\paragraph{Applicability across Training Distributions.}
We further analyze the applicability of \textsc{RL-ARC} across diverse training distributions, with the results presented in Table~\ref{tab:distribution_generalization}. These results demonstrate that \textsc{RL-ARC} consistently improves calibration without substantially sacrificing reasoning performance across both training distributions. Moreover, we observe that \textsc{RL-ARC} robustly improves both calibration and accuracy on OOD benchmarks regardless of the training distribution, compared with other training methods. These results highlight the generalizability of \textsc{RL-ARC} across different training distributions.
\subsection{Reliability on Out-of Distribution (OOD)}
\paragraph{Effectiveness across Tasks.}
To further assess the generalizability of \textsc{RL-ARC} with respect to both accuracy and calibration, we evaluate it on diverse OOD benchmarks. As shown in Table~\ref{tab:generalization_results}, \textsc{RL-ARC} effectively improves calibration on diverse OOD benchmarks while avoiding substantial accuracy degradation. On complex reasoning benchmarks, \textsc{RL-ARC} effectively improves calibration over the existing methods while maintaining competitive accuracy. In particular, on factual QA benchmarks, which are designed to evaluate overconfidence on obscure factual knowledge, \textsc{RL-ARC} achieves the best accuracy and calibration, outperforming the existing calibration-aware training method. These results demonstrate that \textsc{RL-ARC}'s effectiveness in improving calibration while largely preserving reasoning performance generalizes robustly across diverse OOD task types, ranging from tasks that require strong accuracy (i.e., complex reasoning) to those where reliability is critical (i.e., factual QA).

\paragraph{Calibration under Distribution Shift.}
To examine whether the confidence produced by models trained with \textsc{RL-ARC} accurately reflects their correctness given the question, we further analyze confidence-aware calibration on OOD benchmarks that are unseen during training.
As shown in Figure~\ref{fig:main_ece_graph}, \textsc{RL-ARC} estimates confidence that is better aligned with actual prediction correctness than that of the existing calibration-aware approach under distribution shift.
Moreover, when the model is given questions from unseen tasks, \textsc{RL-ARC} assigns more samples to low- and mid-confidence rather than concentrating predictions in high-confidence, which is commonly observed in previous training approaches. 
Together with additional analysis of performance and confidence distributions across ID and OOD settings (see Appendix~\ref{appedix:Confidence-aware Patterns} for detailed results) and calibration gains (see Appendix~\ref{app:additional analysis of calibration gains} for detailed results), these results demonstrate that \textsc{RL-ARC} improves calibration while estimating confidence that better reflects the uncertainty of a given question across both ID and OOD benchmarks.
\begin{figure}[t]
    \centering
    \includegraphics[width=1\linewidth]{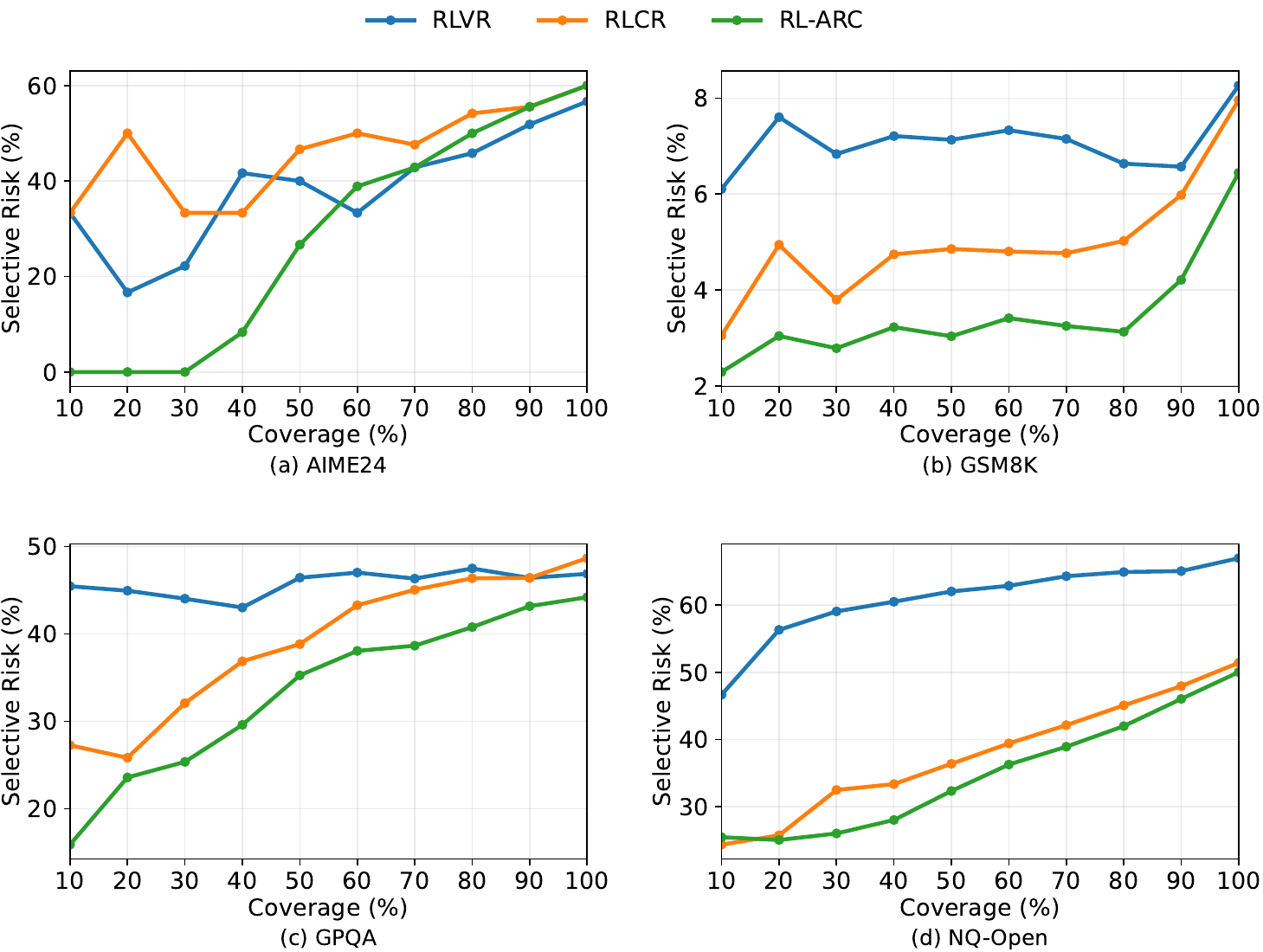}
    \caption{Risk-coverage curves across representative benchmarks. Lower selective risk indicates better confidence estimation and selective prediction performance (i.e., better confidence-aware prediction), where more accurate predictions are assigned higher confidence.}
    \label{fig:aurc_graph}
\end{figure}
\subsection{Overconfidence Mitigation}
If selectively relying on high confidence responses of the model reduces risk, aligning model confidence with output correctness becomes an important practical direction for developing reliable language models. To this end, we further analyze the overconfidence mitigation of \textsc{RL-ARC} through risk-coverage curves, where lower risk at lower coverage indicates that high confidence predictions are more likely to be correct.
As shown in Figure \ref{fig:aurc_graph}, \textsc{RL-ARC} achieves lower or competitive selective risk than existing reasoning training approaches across most coverage levels and benchmarks. Moreover, we observe that \textsc{RL-ARC} further improves both AURC and ECE while achieving competitive accuracy (see Appendix~\ref{appendix:Quantitative Reliability} for a more detailed analysis). Notably, \textsc{RL-ARC} shows substantially lower risk in low coverage regions, where only the model's high-confidence outputs are retained. These results demonstrate that the high confidence outputs produced by \textsc{RL-ARC} are more likely to be correct, suggesting that \textsc{RL-ARC} better aligns confidence with correctness. Compared with existing reasoning training approaches, \textsc{RL-ARC} more effectively filters out incorrect predictions by assigning them lower confidence. These results demonstrate that leveraging reasoning confidence helps mitigate overconfidence on incorrect predictions, highlighting the practical value of \textsc{RL-ARC} in reliable confidence-aware calibration.

\subsection{Component Analysis in \textsc{RL-ARC}}
\paragraph{Ablation Study.}
We conduct an ablation study to analyze the contribution of each auxiliary signal in \textsc{RL-ARC} to both accuracy and calibration. As shown in Table \ref{tab:reward_ablation_ood}, the results demonstrate that incorporating reasoning confidence as an auxiliary signal into calibration-aware training does not lead to a simple one-dimensional improvement. Specifically, the overconfidence penalty, which is applied to incorrect predictions, achieves the best Brier Score and ECE, indicating that applying an additional penalty based on reasoning confidence to incorrect predictions improves calibration while sacrificing accuracy, AUROC, and AURC. Conversely, reasoning-guided regularization, which is applied to correct predictions, serves as a useful signal for improving accuracy, AUROC, and AURC. 
Moreover, we observe that integrating reasoning confidence with an appropriate objective depending on the correctness effectively improves calibration without substantially sacrificing accuracy, compared to alternative strategies for incorporating reasoning confidence (see Appendix~\ref{appendix:integration} for a more detailed analysis). By combining these two objectives, \textsc{RL-ARC} better balances accuracy and calibration, consistently outperforming frameworks that do not leverage reasoning confidence.
\begin{table}[t]
\centering
\setlength{\tabcolsep}{5pt}
\resizebox{\columnwidth}{!}{%
\begin{tabular}{cc|ccccc}
\toprule
\multirow{2}{*}{$\lambda_{\mathrm{neg}}$}
& \multirow{2}{*}{$\lambda_{\mathrm{pos}}$}
& \multicolumn{5}{c}{OOD (Avg.)} \\
\cmidrule(lr){3-7}
&
& Acc. ($\uparrow$)
& AUROC ($\uparrow$)
& AURC ($\downarrow$)
& Brier ($\downarrow$)
& ECE ($\downarrow$) \\
\midrule
\checkmark & \xmark
& 47.11 & 0.53 & \underline{0.46} & \textbf{0.25} & \textbf{0.15} \\

\checkmark & \checkmark
& \underline{47.27} & \underline{0.54} & \textbf{0.45} & \underline{0.26} & \underline{0.16} \\

\xmark & \checkmark
& \textbf{47.51} & \textbf{0.55} & \textbf{0.45} & 0.28 & 0.21 \\

\xmark & \xmark
& 46.02 & \underline{0.54} & 0.47 & 0.28 & 0.20 \\
\bottomrule
\end{tabular}%
}
\caption{
Ablation results on OOD benchmarks.
$\lambda_{\mathrm{neg}}$ and $\lambda_{\mathrm{pos}}$ indicate whether the overconfidence penalty and reasoning-guided regularization are used, respectively. The best and second-best results are highlighted in \textbf{boldface} and \underline{underlined}.
}
\label{tab:reward_ablation_ood}
\end{table}

\begin{figure}[h]
    \centering
    \includegraphics[width=1\linewidth]{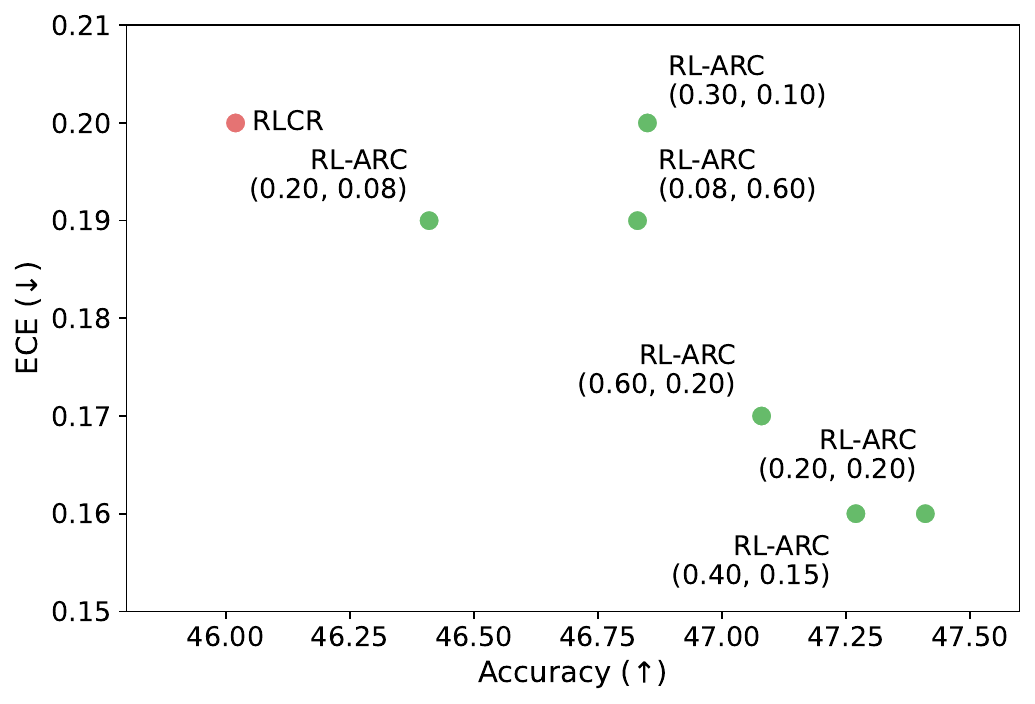}
    \caption{Comparison between performance and calibration across different auxiliary signal scales. For \textsc{RL-ARC}, the left and right values in parentheses denote $\lambda_{\mathrm{pos}}$ and $\lambda_{\mathrm{neg}}$, respectively.
    Here, we use Qwen2.5-7B as base model.
    }
    \label{fig:signal_scales}
\end{figure}
\paragraph{Auxiliary Signal Scale.}
Since the influence of reasoning confidence on calibration in \textsc{RL-ARC} is controlled by the hyperparameters (i.e., $\lambda_{\mathrm{pos}}$ and $\lambda_{\mathrm{neg}}$), we analyze how varying the degree to which reasoning confidence is used affects the effectiveness of \textsc{RL-ARC}.
As shown in Figure~\ref{fig:signal_scales}, \textsc{RL-ARC} consistently outperforms the existing calibration-aware method in calibration while effectively mitigating the accuracy-calibration trade-off. These results demonstrate that reasoning confidence serves as an effective auxiliary signal in \textsc{RL-ARC}.
However, we observe that increasing the influence of reasoning confidence in \textsc{RL-ARC} does not necessarily yield further calibration improvements.
The experimental results suggest that appropriately balancing the auxiliary objectives for correct and incorrect predictions is critical for achieving superior calibration while mitigating the accuracy-calibration trade-off. A more detailed analysis is provided in Appendix~\ref{appendix: Performance Sensitivity to Auxiliary Signal Scales}.

\subsection{Reliability of Reasoning}
\begin{figure}[t]
    \centering
    \includegraphics[width=1\linewidth]{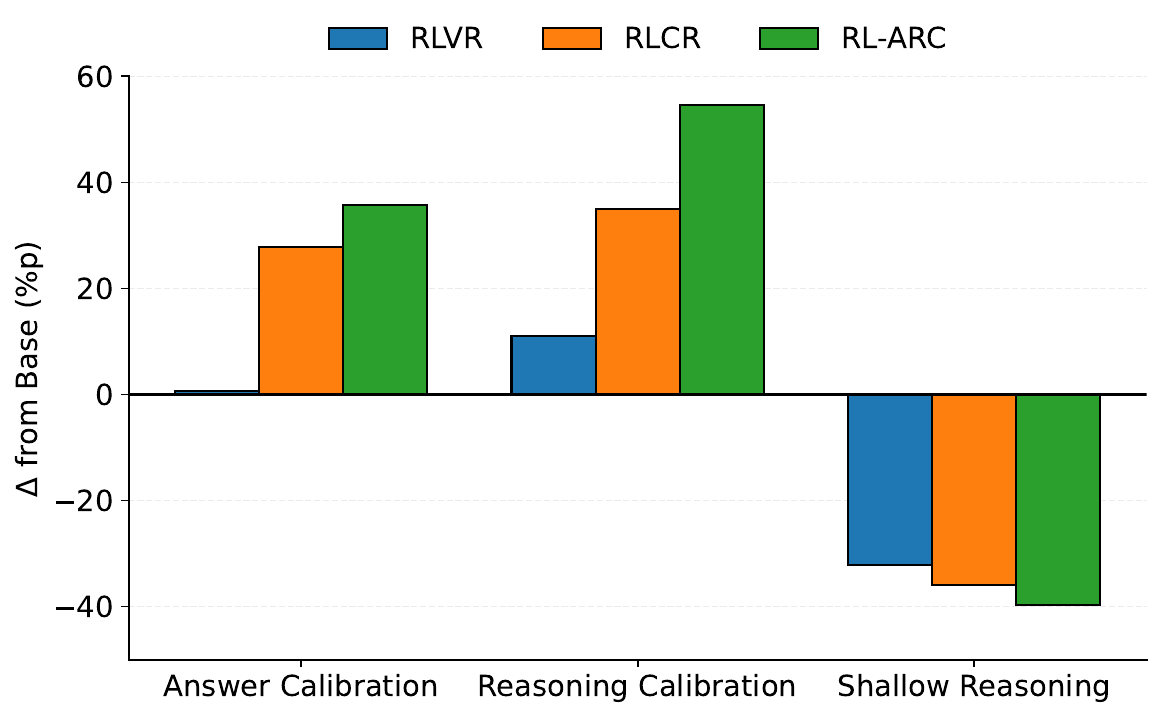}
    \caption{Comparison of calibration improvements and shallow reasoning reduction over the base model. The y-axis reports the relative change from the base model. 
    Here, we use Qwen2.5-7B as base model. 
    }
    \label{fig:reasoning_quality}
\end{figure}

Since reasoning correctness is difficult to assess reliably, \textsc{RL-ARC} leverages reasoning confidence as an auxiliary signal for calibration.
To validate whether \textsc{RL-ARC} improves the reliability of the reasoning process beyond prediction-level calibration, we further analyze reasoning calibration and the frequency of shallow reasoning. As shown in Figure~\ref{fig:reasoning_quality}, compared with existing reasoning training approaches, \textsc{RL-ARC} achieves substantially larger improvements in both answer calibration and reasoning calibration. Notably, compared with the gain in answer calibration, the gain in reasoning calibration is substantially larger than that achieved by the calibration-aware training method. This indicates that explicitly incorporating reasoning confidence into calibration-aware training enables the model to indirectly align its reasoning confidence with reasoning correctness. In addition, \textsc{RL-ARC} achieves the largest reduction in shallow reasoning, indicating that its reasoning process is more strongly aligned with the prediction. These results demonstrate that \textsc{RL-ARC} improves not only the reliability of prediction but also the reliability of the reasoning process itself, even without direct supervision for reasoning correctness, highlighting the effectiveness of leveraging reasoning confidence in calibration-aware reasoning training.

\section{Conclusion}
We have proposed \textsc{RL-ARC}, a simple yet effective calibration-aware training framework that leverages verbalized confidence from both reasoning and prediction. Unlike prior frameworks that have relied solely on confidence in the prediction, \textsc{RL-ARC} leverages confidence in the reasoning process as an auxiliary signal to calibrate confidence in the prediction, enabling reasoning models to improve calibration without substantially sacrificing accuracy. We have extensively evaluated \textsc{RL-ARC} on both in-distribution (ID) benchmarks, such as arithmetic reasoning tasks, and out-of-distribution (OOD) benchmarks, including complex reasoning and factual question answering tasks. The results demonstrate that \textsc{RL-ARC} consistently improves calibration over existing reasoning training approaches, including calibration-aware training methods, across both ID and OOD settings, while achieving accuracy comparable to training methods that focus solely on correctness. Our analysis has validated that, beyond improving calibration, \textsc{RL-ARC} enables reasoning models to adaptively assign confidence based on the given question, thereby highlighting the importance of reasoning confidence in calibration-aware training.

\section*{Limitations}
While \textsc{RL-ARC} effectively improves calibration while mitigating the accuracy-calibration trade-off, \textsc{RL-ARC} has several limitations that future work could address:
First, our experiments are limited to two training settings, in which LMs are trained on either a limited set of mathematical reasoning samples or a limited set of complex reasoning samples. Additionally, to ensure a fair comparison with the baselines, we train all LMs using only GRPO.
We believe that incorporating more diverse training datasets and stronger RL algorithms could further improve model performance.
Second, we mainly focus on LMs, particularly the Qwen family, which is widely used in RL tasks. We believe that extending the \textsc{RL-ARC} framework to calibrate multimodal reasoning models presents a promising avenue for future work. Finally, to compute reasoning calibration and the frequency of shallow reasoning, we utilize LLM-as-a-judge. However, we observe that using LLM-as-a-judge to assess the correctness of the reasoning process can introduce errors. We leave the development of more accurate LLM-based evaluation methods for reasoning correctness to future work.

\section*{Acknowledgments}
This work was supported by the National Research Foundation of Korea (NRF) grant funded by the Korea government (MSIT) (No.RS-2025-00517221 and No.RS-2024-00415812) and Institute of Information \& communications Technology Planning \& Evaluation (IITP) grant funded by the Korea government (MSIT) (No.RS-2024-00439328, Karma: Towards Knowledge Augmentation for Complex Reasoning (SW Starlab), No.RS-2024-00457882, AI Research Hub Project, and No.RS-2019-II190079, Artificial Intelligence Graduate School Program (Korea University)).



\newpage
\bibliography{custom}
\clearpage
\appendix
\section{Experimental Details}
\label{appendix:experiment_details}
\subsection{Dataset Details}
\label{appendix;dataset_details}
We provide a brief overview of each dataset along with key statistics below.\\
\textbf{Big-Math}~\citep{albalak2025bigmathlargescalehighqualitymath} is a large-scale training dataset for mathematical reasoning, comprising 250{,}000 math problems from a diverse set of benchmarks, including MATH~\citep{hendrycks2021measuring} and GSM8K~\citep{cobbe2021trainingverifierssolvemath}. For training, following the training setup of \citep{damani2026beyond}, we utilize 30{,}000 problems that fall within an appropriate difficulty range. For evaluation, we use 1{,}000 problems and verify the correctness of model predictions using \texttt{math-verify}\footnote{https://github.com/huggingface/Math-Verify.}.\\
\textbf{GSM8K} \citep{cobbe2021trainingverifierssolvemath} is a representative generative dataset to assess a model’s arithmetic reasoning ability on high-quality and linguistically diverse grade school math word problems.
For evaluation, we use 1{,}319 problems and assess the model's predictions with \texttt{math-verify}.\\
\textbf{MATH-500} \citep{hendrycks2021measuring} is a widely used dataset for evaluating mathematical reasoning capability, consisting of a subset of 500 problems from the MATH benchmark~\cite{hendrycks2021measuring}. We verify the model predictions using \texttt{math-verify}.\\
\textbf{AMC23}\footnote{https://huggingface.co/datasets/math-ai/amc23} is a representative benchmark for evaluating the arithmetic reasoning ability of LMs, with an emphasis on functional equations and complex reasoning. This benchmark consists of 40 problems from the 2023 American Mathematics Competition. For automatic evaluation, we use \texttt{math-verify}.\\
\textbf{AIME24}~\cite{aime24} is a widely used mathematical reasoning benchmark, consisting of 30 challenging problems from the 2024 American Invitational Mathematics Examination. We verify the model predictions using \texttt{math-verify}.\\
\textbf{AIME25}~\cite{aime25} is a widely used mathematical reasoning dataset, consisting of 30 challenging problems from the 2025 American Invitational Mathematics Examination. We assess the model's predictions using \texttt{math-verify}.\\
\textbf{StrategyQA} \citep{geva-etal-2021-aristotle} is a representative complex reasoning dataset designed to evaluate models' ability to perform implicit multi-hop reasoning. We evaluate the model on 229 questions using LLM-as-a-judge~\cite{zheng2023judging}.\\
\textbf{GPQA} \citep{rein2024gpqa} is a widely used benchmark to evaluate the model's complex reasoning ability. For evaluation, we use 448 questions from the main dataset, created by experts in biology, chemistry, and physics. We assess the model's performance using LLM-as-a-judge.\\
\textbf{HotpotQA} \citep{yang-etal-2018-hotpotqa} is a representative dataset designed to evaluate models' ability to identify and reason over relevant facts across multiple documents. For evaluation, following the experimental setup of \citep{damani2026beyond}, we use a subsampled dataset consisting of 1{,}000 questions, where two non-relevant paragraphs are removed from each question. We evaluate the model's reasoning ability using exact match.\\
\textbf{SimpleQA} \citep{wei2024measuringshortformfactualitylarge} is a representative benchmark designed to evaluate the model's ability to answer fact-seeking questions. We assess model performance on 4{,}326 factual questions using LLM-as-a-judge.\\
\textbf{Natural Questions (NQ-Open)} \citep{kwiatkowski-etal-2019-natural} is a widely used factual question answering dataset constructed from Google Search queries paired with annotated short answers. We evaluate the model's performance on 3{,}610 questions using LLM-as-a-judge.\\
\textbf{TriviaQA} \citep{joshi-etal-2017-triviaqa} is a widely used factual question answering dataset for evaluating models' factual knowledge. We evaluate the model's factual reasoning ability on a subsampled dataset of 2{,}000 questions in a closed-book setting. We measure the model's reasoning performance using LLM-as-a-judge.

\begin{table}[ht]
\centering
\begin{tabular}{lc}
\hline
Hyperparameter           & Value        \\ \hline
Maximum input length     & 1024          \\
Maximum output length     & 4096          \\
Rollout group size       & 8          \\
Rollout temperature       & 0.7          \\
Epoch                    & 1            \\
Batch size               & 1536           \\
Learning rate            & 5e-6         \\
Learning rate scheduling & Linear decay \\
Warmup ratio             & 0.2          \\ \hline
\end{tabular}
\caption{Hyperparameter settings for training.
}
\label{table: hyperparameter setting}
\end{table}
\subsection{Implementation}
\label{appendix:implmentation}
Following the training setup of \cite{damani2026beyond}, we use Big-Math~\cite{albalak2025bigmathlargescalehighqualitymath} as the training dataset, excluding questions that LLaMA-8B can solve easily and retaining only those with numerical answers. We conduct all training on four NVIDIA H100 (or A100 GPUs). To ensure a fair comparison, all training methods use the same hyperparameters. In our work, we set the default values of $\lambda_{\mathrm{pos}}$ and $\lambda_{\mathrm{neg}}$ in \textsc{RL-ARC} to 0.4 and 0.15, respectively. The detailed hyperparameter settings used in our work are described in Table~\ref{table: hyperparameter setting}. 

\subsection{Metrics Details}
\label{appendix:metrics_details}
\textbf{Accuracy} is a metric for measuring the model's reasoning performance.\\ 
\textbf{Brier Score (Brier)} \citep{Brier1950VERIFICATIONOF} is a widely used proper scoring rule for assessing the accuracy of a model's probabilistic predictions. In our work, it is measured as the squared difference between the verbalized confidence and correctness, and is defined as follows:
\begin{equation}
\mathrm{Brier}(y, c_a, y^{*})
=
\left(\mathbf{1}_{y=y^*} - c_a\right)^{2}.
\end{equation}
Here, $y$ is the prediction generated by the LMs $\pi_{\theta}$, $y^*$ is the ground-truth answer, $\mathbf{1}_{y=y^*}$ is the indicator function that evaluates whether $y$ is correct to $y^*$, and $c_a$ is the self-assessed confidence of the LM $\pi_{\theta}$ associated with $y$.\\
\textbf{Expected Calibration Error (ECE)} is a representative calibration metric, as defined below:
\begin{equation}
\mathrm{ECE} = \sum_{g=1}^{G} \frac{|B_g|}{N} \left| \mathrm{acc.}(B_g) - \mathrm{conf.}(B_g) \right|.
\end{equation}
Here, $G$ denotes the number of bins, $B_g$ denotes the set of samples assigned to bin \(g\), and $N$ denotes the total number of samples. \(\mathrm{acc.}(B_g)\) and \(\mathrm{conf.}(B_g)\) represent the average accuracy and average confidence of the samples in bin \(g\), respectively. In this work, we measure the model's calibration using ECE with \(G=10\) bins.\\
\textbf{Area Under the ROC Curve (AUROC)} is a widely used metric for measuring how well confidence scores distinguish correct predictions from incorrect ones, as defined below:
\begin{equation}
\mathrm{AUROC} = \int_{0}^{1} \mathrm{TPR}\left(\mathrm{FPR}^{-1}(t)\right) \, dt.
\end{equation}
Here, \(\mathrm{TPR}\) denotes the true positive rate and \(\mathrm{FPR}\) denotes the false positive rate.\\
\textbf{Area Under the Risk-Coverage Curve (AURC)} is a selective prediction metric that measures how risk changes as coverage varies, as defined below:
\begin{equation}
\mathrm{AURC}
=
\int_{0}^{1} R(c)\,dc,
\quad
R(c)
=
1 - \mathrm{acc.}(T_c).
\end{equation}
Here, \(c\) denotes the fraction of selected samples, \(R(c)\) denotes the selective risk at coverage \(c\), and \(T_c\) denotes the top-\(c\) fraction of samples sorted by confidence. \(\mathrm{acc}(T_c)\) denotes the accuracy computed over the selected subset \(T_c\).\\
\textbf{Reasoning Calibration} is a metric for measuring how well the model's reasoning confidence \(c_r\) reflects the correctness of its reasoning process. Specifically, we first determine reasoning correctness by assessing whether the model's reasoning process supports the ground-truth answer using LLM-as-a-judge~\citep{zheng2023judging, Wang_Tu_Wang_Wu_Liu_Mei_Bartoldson_Kailkhura_Xie_2026}. We then group samples into bins according to reasoning confidence \(c_r\) and compute the gap between the average reasoning correctness and the average reasoning confidence within each bin (i.e., ECE).\\
\textbf{Shallow Reasoning (SR)} is a metric for measuring how often a reasoning model produces correct predictions while generating incorrect or irrelevant reasoning processes. We first evaluate both prediction correctness and reasoning correctness for the given question. We then compute the proportion of cases where the model produces a correct prediction but an incorrect reasoning process on OOD benchmarks, as follows:
\begin{equation}
\mathrm{SR}
=
\frac{1}{N}
\sum_{i=1}^{N}
\mathbb{I}
\left[
\hat{y}_i = y_i
\;\land\;
r_i = 0
\right].
\end{equation}
Here, \(N\) is the total number of samples, \(\hat{y}_i\) and \(y_i\) are the model prediction and the ground-truth answer for the \(i\)-th sample, respectively, and \(r_i\) is reasoning correctness.

\subsection{Calibration-aware Reasoning Training Prompt Template}
\label{appendix:Calibration-aware Reasoning Training Prompt Template}
\begin{figure*}[h]
    \centering
    \includegraphics[width=1\linewidth]{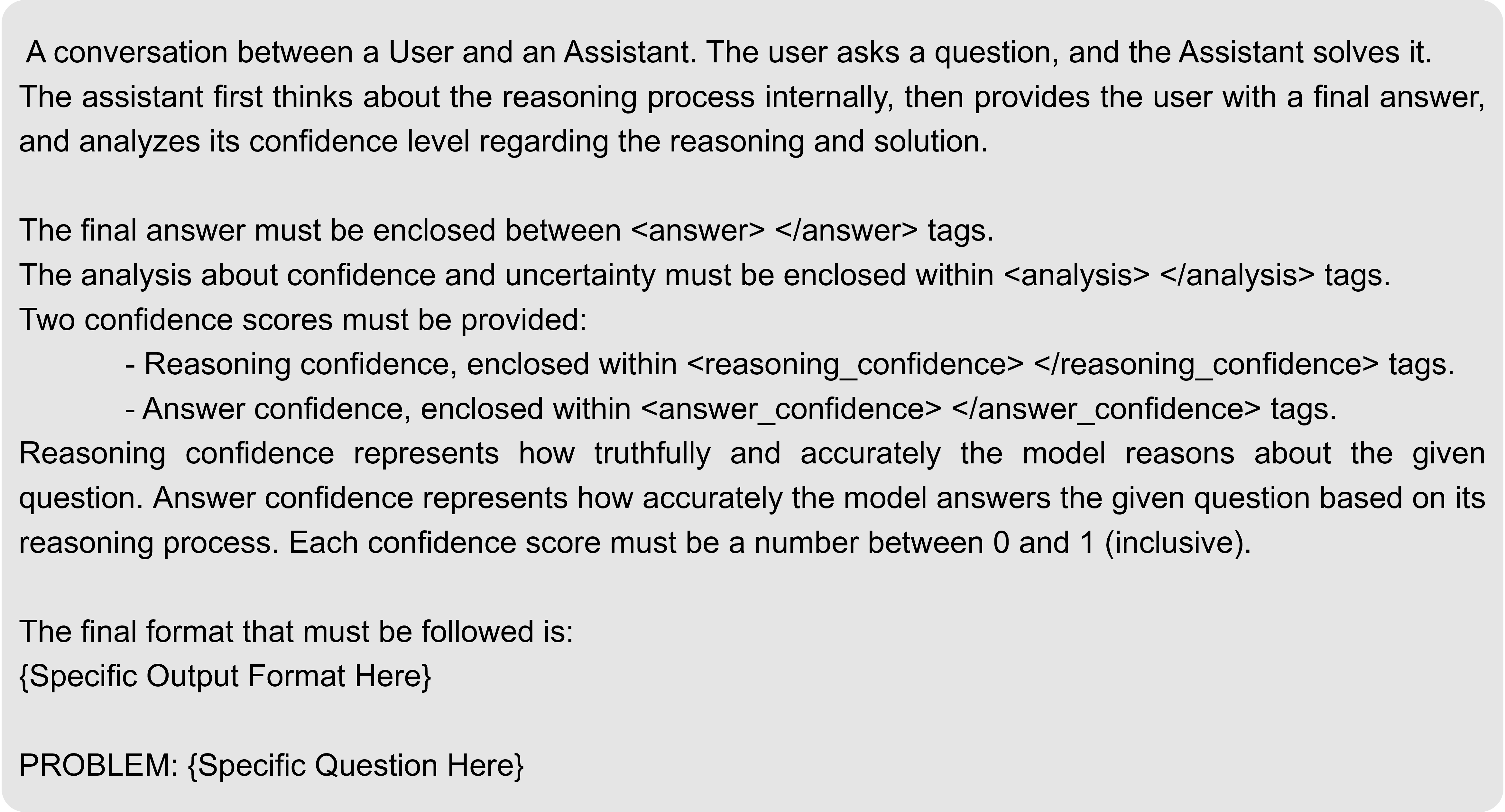}
    \caption{The designed prompt used to estimate verbalized confidence during reasoning.}
    \label{fig:prompt_template}
\end{figure*}

We design a prompt template for calibration-aware reasoning training that leverages confidence in both the reasoning process and predictions. The detailed template is illustrated in Figure~\ref{fig:prompt_template}.

\begin{figure*}[ht]
    \centering
    \includegraphics[width=1\linewidth]{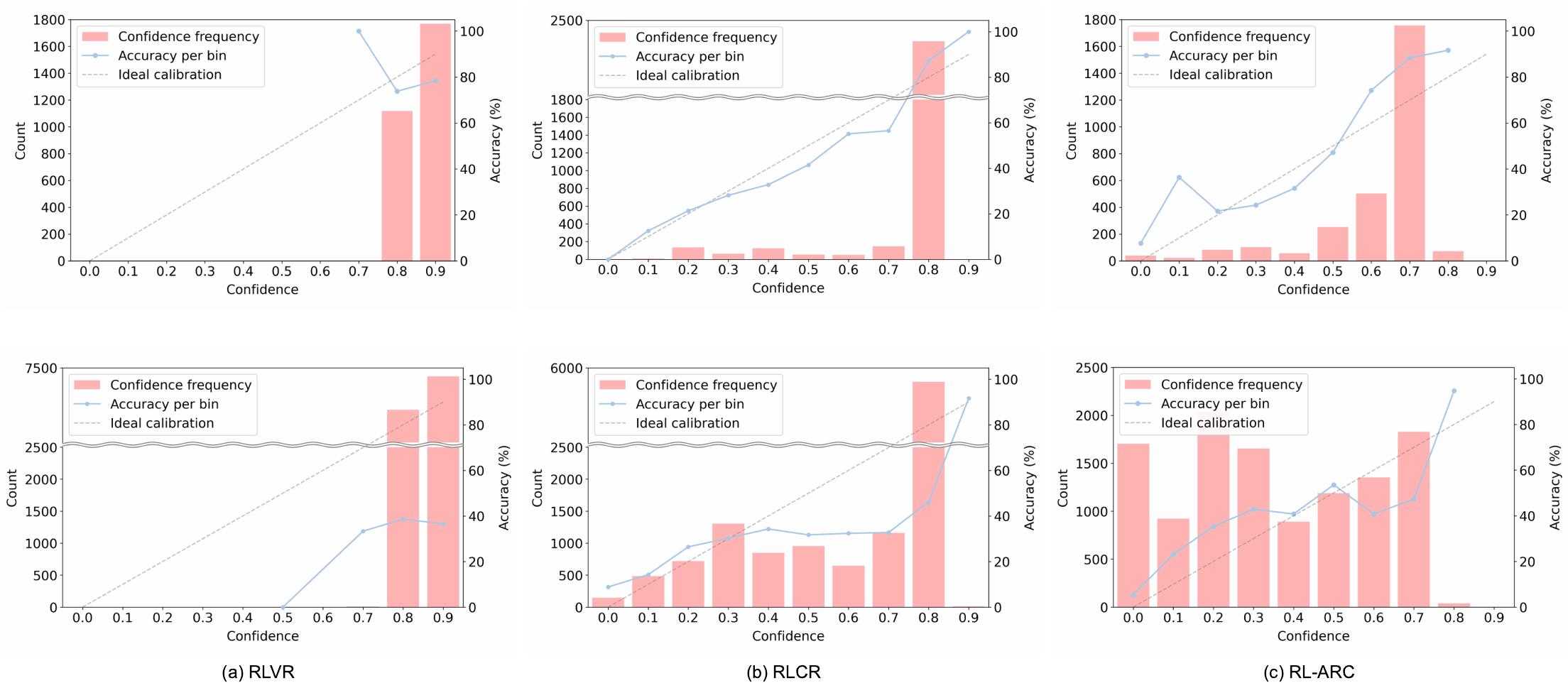}
    \caption{Performance (solid lines) and confidence frequency (bars) across confidence bins for Qwen2.5-7B trained with each method on ID (top) and OOD (bottom) benchmarks.}
    \label{fig:app_ece_qwen25}
\end{figure*}
\begin{figure*}[t]
    \centering
    \includegraphics[width=1\linewidth]{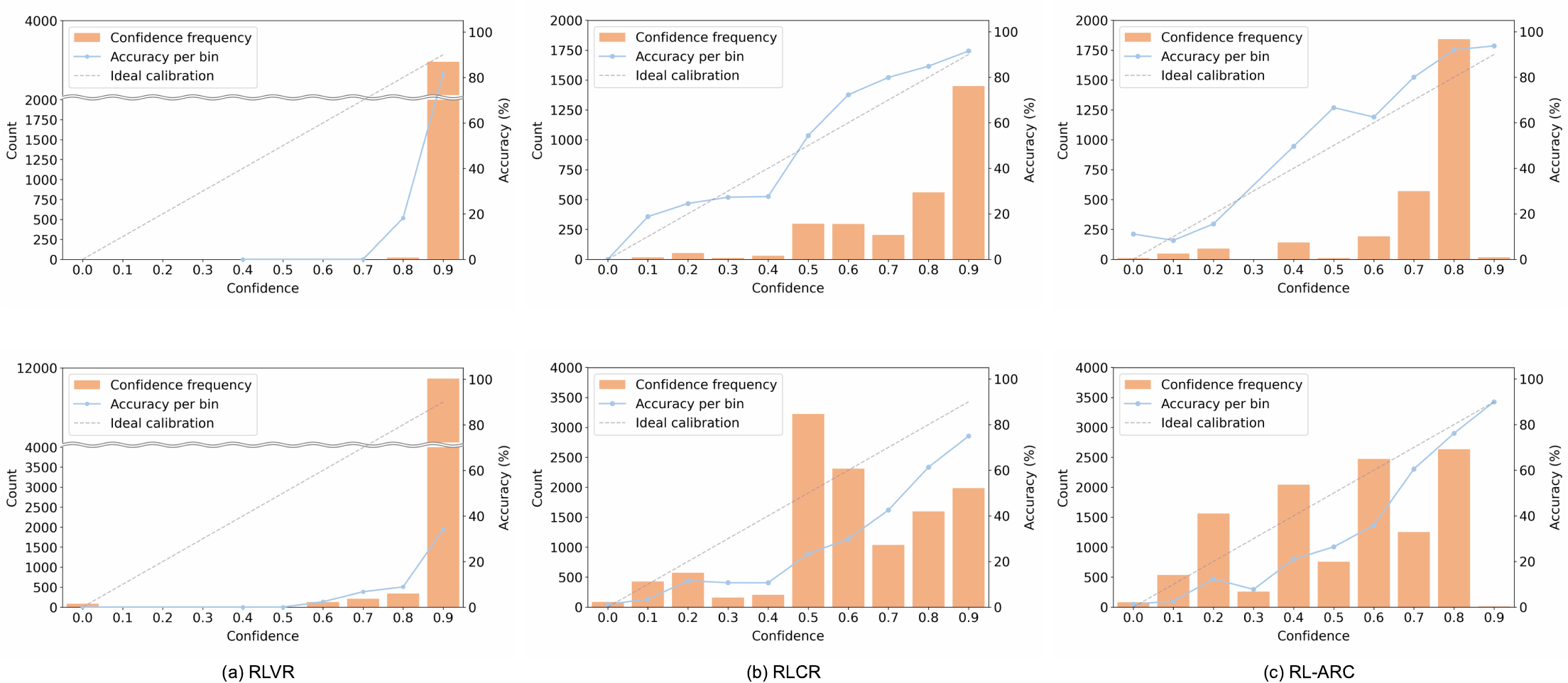}
    \caption{Performance (solid lines) and confidence frequency (bars) across confidence bins for Qwen3-8B trained with each method on ID (top) and OOD (bottom) benchmarks.}
    \label{fig:app_ece_qwen3}
\end{figure*}

\section{Detailed Experiment Results}
\subsection{Overall Results for Each Task}
\label{appendix:overall_results_for_each_task}
We report the overall results for each benchmark across all methods considered in this work, including \textsc{RL-ARC}. The detailed results are provided below:
\begin{itemize}
    \item \textbf{Qwen2.5-7B (Big-Math)}: The results on math benchmarks are reported in Table~\ref{tab:appendix: qwen2.5 overall-resutls id}; the results on complex reasoning benchmarks are reported in Table~\ref{tab:appendix: qwen2.5 overall-resutls ood complex reasoning}; and the results on factual question answering benchmarks are reported in Table~\ref{tab:appendix: qwen2.5 overall-resutls ood factual}.
    \item \textbf{Qwen2.5-7B (HotpotQA)}: The results on complex reasoning benchmarks are reported in Table~\ref{tab:appendix:qwenhotpot-overall-results-ood-complex}; and the results on factual question answering benchmarks are reported in Table~\ref{tab:appendix:qwen_hotpotqa_overall-results-ood-factual}.
    \item \textbf{Qwen3-8B}: The results on math benchmarks are reported in Tables~\ref{tab:appendix: qwen3 overall-resutls id} and \ref{tab:appendix: qwen3 overall-resutls id-part2}; the results on complex reasoning benchmarks are reported in Table~\ref{tab:appendix: qwen3 overall-resutls ood complex resonign}; and the results on factual question answering benchmarks are reported in Table~\ref{tab:appendix: qwen3 overall-resutls ood factual}.
    \item \textbf{Llama-8B (DeepSeek-R1 Distilled)}: The results on math benchmarks are reported in Table~\ref{tab:appendix:deepseekllama-overall-results-id}; the results on complex reasoning benchmarks are reported in Table~\ref{tab:appendix:deepseekllama-overall-results-ood-complex}; and the results on factual question answering benchmarks are reported in Table~\ref{tab:appendix:deepseekllama-overall-results-ood-factual}.
\end{itemize}

\subsection{Confidence-aware Patterns}
\label{appedix:Confidence-aware Patterns}
As shown in Figures~\ref{fig:app_ece_qwen25} and \ref{fig:app_ece_qwen3}, in ID settings, \textsc{RL-ARC} shows confidence allocation patterns comparable to RLCR, indicating that both calibration-aware methods can improve confidence behavior over standard RLVR. However, the advantage of \textsc{RL-ARC} becomes more pronounced under OOD settings. While RLVR and RLCR still concentrate many predictions in high-confidence regions, \textsc{RL-ARC} distributes predictions over a broader range of confidence levels and assigns low or mid confidence to uncertain predictions. This suggests that incorporating reasoning confidence helps preserve confidence-aware behavior under distribution shift.

\begin{table}[t]
\centering
\setlength{\tabcolsep}{7pt}
\resizebox{\columnwidth}{!}{
\renewcommand{\arraystretch}{1.1}
\begin{tabular}{lcc|cc}
\toprule
\multirow{2}{*}{Method}
& \multicolumn{2}{c|}{ID (Avg.)}
& \multicolumn{2}{c}{OOD (Avg.)} \\
\cmidrule(lr){2-3}
\cmidrule(lr){4-5}
& Brier ($\downarrow$)
& ECE ($\downarrow$)
& Brier ($\downarrow$)
& ECE ($\downarrow$) \\
\midrule
RLVR
& 0.41 & 0.40
& 0.50 & 0.50 \\

RLCR
& 0.25 & 0.22
& 0.28 & 0.20 \\

\textsc{RL-ARC} (shuffled)
& 0.25 & 0.19
& 0.27 & 0.19 \\

\cellcolor[gray]{.9}\textsc{RL-ARC}
& \cellcolor[gray]{.9}0.23 & \cellcolor[gray]{.9}0.18
& \cellcolor[gray]{.9}0.26 & \cellcolor[gray]{.9}0.16 \\
\bottomrule
\end{tabular}
}
\caption{
Control experiment results on the calibration gains of \textsc{RL-ARC} across ID and OOD benchmarks.}
\label{tab:shuffled_confidence}
\end{table}

\subsection{Additional Analysis of Calibration Gains}
\label{app:additional analysis of calibration gains}
Since lower calibration error can result either from better confidence-accuracy alignment within each confidence bin or simply from a flatter confidence distribution, we analyze whether models trained with \textsc{RL-ARC} achieve improved calibration within individual confidence bins. To disentangle confidence-accuracy alignment from changes in the confidence distribution, we conduct a control experiment in which \textsc{RL-ARC}’s confidence scores were randomly shuffled across samples while preserving the exact confidence histogram and bin occupancy. As shown in Table~\ref{tab:shuffled_confidence}, \textsc{RL-ARC} consistently achieves lower ECE and Brier than its shuffled counterpart on both ID and OOD benchmarks. These results demonstrate that the calibration gains cannot be attributed solely to a broader or flatter confidence distribution.

\subsection{Quantitative Reliability}
\label{appendix:Quantitative Reliability}
Since AURC primarily captures the quality of confidence-based ranking under selective prediction, we further analyze this tendency using AURC together with accuracy and ECE. As shown in Table~\ref{tab:id_ood_calibration}, \textsc{RL-ARC} achieves the lowest AURC in both ID and OOD settings, indicating better selective prediction performance across diverse task types. Notably, compared with RLCR, we observe that \textsc{RL-ARC} further improves both AURC and ECE while maintaining competitive accuracy in ID settings and achieving the best accuracy in OOD settings. Overall, these results demonstrate that \textsc{RL-ARC} not only improves calibration and performance but also enhances the practical reliability of confidence-aware prediction.
\begin{table}[h]
\centering
\setlength{\tabcolsep}{2.5pt}
\resizebox{\columnwidth}{!}{
\begin{tabular}{lccc|ccc}
\toprule

\multirow{2}{*}{Method}
& \multicolumn{3}{c|}{ID (Avg.)}
& \multicolumn{3}{c}{OOD (Avg.)} \\

\cmidrule(lr){2-4}
\cmidrule(lr){5-7}
& Acc.\,($\uparrow$)
& AURC\,($\downarrow$)
& ECE\,($\downarrow$)
& Acc.\,($\uparrow$)
& AURC\,($\downarrow$)
& ECE\,($\downarrow$) \\
\midrule

RLVR
& 66.58 & 0.22 & 0.32
& 47.24 & 0.44 & 0.48 \\

RLCR
& 65.98 & 0.20 & 0.16
& 50.99 & 0.35 & 0.22 \\

\cellcolor[gray]{.9}\textsc{RL-ARC}
& \cellcolor[gray]{.9}66.10 & \cellcolor[gray]{.9}0.16 & \cellcolor[gray]{.9}0.15
& \cellcolor[gray]{.9}51.61 & \cellcolor[gray]{.9}0.34 & \cellcolor[gray]{.9}0.17 \\

\bottomrule
\end{tabular}
}
\caption{
Comparison of accuracy (\%), AURC, and ECE under Math (ID) and OOD benchmarks.}
\label{tab:id_ood_calibration}
\end{table}

\begin{table*}[ht]
\centering
\resizebox{\textwidth}{!}{
\begin{tabular}{lcccccccc}
\toprule
\multirow{2}{*}{Method}
& \multicolumn{4}{c}{ID (Avg.)}
& \multicolumn{4}{c}{OOD (Avg.)} \\
\cmidrule(lr){2-5}
\cmidrule(lr){6-9}
& Acc.\,($\uparrow$) & AUROC\,($\uparrow$) & Brier\,($\downarrow$) & \cellcolor{teal!10}ECE\,($\downarrow$)
& Acc.\,($\uparrow$) & AUROC\,($\uparrow$) & Brier\,($\downarrow$) & \cellcolor{teal!10}ECE\,($\downarrow$) \\
\midrule
RLVR & 56.86 & 0.45 & 0.41 & \cellcolor{teal!10}0.40 & 47.58 & 0.50 & 0.50 & \cellcolor{teal!10}0.50 \\\midrule
\multicolumn{1}{l}{\textsc{RL-ARC}} \\        
w/ Arithmetic mean & 53.42 & 0.58 & 0.27 & \cellcolor{teal!10}0.24 & 48.61 & 0.54 & 0.31 & \cellcolor{teal!10}0.25 \\
w/ Geometric mean & 54.04  & 0.58  & 0.26  & \cellcolor{teal!10}0.25  & 47.65  & 0.53  & 0.30  & \cellcolor{teal!10}0.23  \\
\cellcolor[gray]{.9}w/ Ours & \cellcolor[gray]{.9}56.53 & \cellcolor[gray]{.9}0.60  & \cellcolor[gray]{.9}0.23  & \cellcolor{teal!10}0.18  & \cellcolor[gray]{.9}47.27 & \cellcolor[gray]{.9}0.54  & \cellcolor[gray]{.9}0.26  & \cellcolor{teal!10}0.16  \\
\bottomrule
\end{tabular}
}
\caption{Comparison of strategies for integrating reasoning confidence into calibration-aware training on Math (ID) and OOD benchmarks.}
\label{tab:reasoning-confidence-integration}
\end{table*}

\subsection{Performance Sensitivity to Reasoning Confidence Integration}
\label{appendix:integration}
Since leveraging reasoning confidence as an auxiliary signal is a key component of \textsc{RL-ARC}, we further analyze how effectively our integration strategy improves calibration while minimizing performance degradation compared with alternative strategies. Specifically, we compare our integration strategy with alternative integration strategies in terms of accuracy and calibration, including the arithmetic mean and geometric mean as representative weighted-sum and weighted-product approaches, respectively \citep{he-etal-2025-mmboundary, lee2026knowproxy}. As shown in Table~\ref{tab:reasoning-confidence-integration}, we observe that our strategy for integrating reasoning confidence, which employs distinct objectives depending on correctness, effectively improves calibration while mitigating the accuracy–calibration trade-off compared to alternative strategies. These results suggest that using reasoning confidence for reasoning-guided regularization in correct cases while imposing an additional overconfidence penalty in incorrect cases is crucial for effectively mitigating the accuracy–calibration trade-off in \textsc{RL-ARC}.

\begin{table*}[ht]
\centering
\small
\setlength{\tabcolsep}{4pt}
\resizebox{\textwidth}{!}{
\begin{tabular}{cc|cccc|cccc}
\toprule
\multicolumn{2}{c|}{Setting}
& \multicolumn{4}{c|}{ID (Avg.)}
& \multicolumn{4}{c}{OOD (Avg.)} \\
\cmidrule(lr){1-2}
\cmidrule(lr){3-6}
\cmidrule(lr){7-10}
\multicolumn{1}{c}{$\lambda_{\mathrm{neg}}$}
& \multicolumn{1}{c|}{$\lambda_{\mathrm{pos}}$}
& \multicolumn{1}{c}{Acc.\,($\uparrow$)}
& \multicolumn{1}{c}{AUROC\,($\uparrow$)}
& \multicolumn{1}{c}{Brier\,($\downarrow$)}
& \multicolumn{1}{c|}{\cellcolor{teal!10}ECE\,($\downarrow$)}
& \multicolumn{1}{c}{Acc.\,($\uparrow$)}
& \multicolumn{1}{c}{AUROC\,($\uparrow$)}
& \multicolumn{1}{c}{Brier\,($\downarrow$)}
& \multicolumn{1}{c}{\cellcolor{teal!10}ECE\,($\downarrow$)} \\
\midrule

\multicolumn{1}{c}{0.00} & \multicolumn{1}{c|}{0.00} & 54.82 & 0.58 & 0.25 & \cellcolor{teal!10}0.22 & 46.02 & 0.54 & 0.28 & \cellcolor{teal!10}0.20 \\
\multicolumn{1}{c}{0.08} & \multicolumn{1}{c|}{0.20} & 55.92 & 0.61 & 0.24 & \cellcolor{teal!10}0.20 & 46.41 & 0.55 & 0.28 & \cellcolor{teal!10}0.19 \\
\multicolumn{1}{c}{0.10} & \multicolumn{1}{c|}{0.30} & 54.11 & 0.64 & 0.24 & \cellcolor{teal!10}0.21 & 46.85 & 0.54 & 0.28 & \cellcolor{teal!10}0.20 \\
\multicolumn{1}{c}{0.15} & \multicolumn{1}{c|}{0.40} & 56.53 & 0.60 & \textbf{0.23} & \cellcolor{teal!10}\textbf{0.18} & 47.27 & 0.54 & \textbf{0.26} & \cellcolor{teal!10}\textbf{0.16} \\
\multicolumn{1}{c}{0.20} & \multicolumn{1}{c|}{0.60} & \textbf{57.64} & \textbf{0.65} & \textbf{0.23} & \cellcolor{teal!10}0.19 & 47.08 & \textbf{0.56} & 0.27 & \cellcolor{teal!10}0.17 \\
\multicolumn{1}{c}{0.20} & \multicolumn{1}{c|}{0.20} & 54.30 & 0.62 & 0.25 & \cellcolor{teal!10}0.20 & \textbf{47.41} & 0.54 & \textbf{0.26} & \cellcolor{teal!10}\textbf{0.16} \\
\multicolumn{1}{c}{0.60} & \multicolumn{1}{c|}{0.08} & 55.97 & \textbf{0.65} & 0.24 & \cellcolor{teal!10}0.20 & 46.83 & 0.54 & 0.28 & \cellcolor{teal!10}0.19 \\
\bottomrule
\end{tabular}
}
\caption{Detailed evaluation results of auxiliary signal scaling for \textsc{RL-ARC} across ID and OOD benchmarks. The best results are highlighted in \textbf{boldface}.}
\label{tab:auxiliary-signal-scale}
\end{table*}

\subsection{Performance Sensitivity to Auxiliary Signal Scales}
\label{appendix: Performance Sensitivity to Auxiliary Signal Scales}
We further analyze the sensitivity of \textsc{RL-ARC}'s performance to different auxiliary signal scales. As shown in Table~\ref{tab:auxiliary-signal-scale}, \textsc{RL-ARC} consistently achieves a better accuracy-calibration trade-off across a wide range of hyperparameter configurations. Moreover, we observe that the optimal hyperparameter configurations for effectively mitigating the accuracy–calibration trade-off are predominantly positive-dominant (i.e., $\lambda_{\mathrm{pos}} / \lambda_{\mathrm{neg}} > 1$), rather than balanced (i.e., $\lambda_{\mathrm{pos}} / \lambda_{\mathrm{neg}} = 1$) or negative-dominant (i.e., $0 < \lambda_{\mathrm{pos}} / \lambda_{\mathrm{neg}} < 1$). Notably, taken together with the results in Table~\ref{tab:reward_ablation_ood}, we observe that the positive-case regularizer primarily affects accuracy and AUROC, whereas the negative-case overconfidence penalty mainly improves calibration, particularly in the positive-dominant settings.

\begin{table*}[ht]
\tiny
\centering
\resizebox{\textwidth}{!}{
\begin{tabular}{llcccc}
\toprule
Benchmark & Method
& Acc.\,($\uparrow$) 
& AUROC\,($\uparrow$) 
& Brier\,($\downarrow$) 
& \cellcolor{teal!10}ECE\,($\downarrow$) \\
\midrule

\multirow{7}{*}{Big-Math}
& Base & 0.50 & 0.55 & 0.46 & \cellcolor{teal!10}0.46 \\
& \multicolumn{1}{l}{RLVR} \\        
& \hspace{0.4em}w/ Confidence & \textbf{0.67} & 0.51 & 0.32 & \cellcolor{teal!10}0.32 \\
& \hspace{0.4em}w/ Probability & \textbf{0.67} & 0.61 & 0.31 & \cellcolor{teal!10}0.31 \\
& \hspace{0.4em}w/ Post-hoc & \textbf{0.67} & \textbf{0.73} & \textbf{0.21 }& \cellcolor{teal!10}0.08 \\
& Behavioral Calibration & \textbf{0.67} & 0.50 & 0.22 & \cellcolor{teal!10}\textbf{0.03} \\
& RLCR & 0.66 & 0.59 & 0.22 & \cellcolor{teal!10}\textbf{0.03} \\
& \cellcolor[gray]{.9}\textsc{RL-ARC} (ours) & \cellcolor[gray]{.9}\textbf{0.67} & \cellcolor[gray]{.9}0.60 & \cellcolor[gray]{.9}\textbf{0.21} & \cellcolor{teal!10}0.06 \\

\midrule

\multirow{7}{*}{GSM8K}
& Base & 0.73 & 0.52 & 0.25 & \cellcolor{teal!10}0.22 \\
& \multicolumn{1}{l}{RLVR} \\        
& \hspace{0.4em}w/ Confidence & 0.88 & 0.48 & 0.12 & \cellcolor{teal!10}\textbf{0.10} \\
& \hspace{0.4em}w/ Probability & 0.88 & 0.56 & \textbf{0.11} & \cellcolor{teal!10}\textbf{0.10} \\
& \hspace{0.4em}w/ Post-hoc & 0.88 & \textbf{0.66} & 0.16 & \cellcolor{teal!10}0.20 \\
& Behavioral Calibration & 0.89 & 0.48 & 0.14 & \cellcolor{teal!10}0.20 \\
& RLCR & \textbf{0.90} & 0.64 & 0.15 & \cellcolor{teal!10}0.24 \\
& \cellcolor[gray]{.9}\textsc{RL-ARC} (ours) & \cellcolor[gray]{.9}0.89 & \cellcolor[gray]{.9}0.57 & \cellcolor[gray]{.9}0.16 & \cellcolor{teal!10}0.25 \\

\midrule

\multirow{7}{*}{MATH-500}
& Base & 0.45 & 0.58 & 0.50 & \cellcolor{teal!10}0.51 \\
& \multicolumn{1}{l}{RLVR} \\        
& \hspace{0.4em}w/ Confidence & \textbf{0.64} & 0.49 & 0.35 & \cellcolor{teal!10}0.35 \\
& \hspace{0.4em}w/ Probability & \textbf{0.64} & 0.62 & 0.34 & \cellcolor{teal!10}0.34 \\
& \hspace{0.4em}w/ Post-hoc & \textbf{0.64} & \textbf{0.73} & \textbf{0.20} & \cellcolor{teal!10}0.12 \\
& Behavioral Calibration & 0.55 & 0.50 & 0.27 & \cellcolor{teal!10}0.15 \\
& RLCR & 0.57 & 0.52 & 0.26 & \cellcolor{teal!10}0.10 \\
& \cellcolor[gray]{.9}\textsc{RL-ARC} (ours) & \cellcolor[gray]{.9}0.58 & \cellcolor[gray]{.9}0.57 & \cellcolor[gray]{.9}0.24 & \cellcolor{teal!10}\textbf{0.06} \\

\midrule

\multirow{7}{*}{AMC23}
& Base & 0.35 & 0.49 & 0.60 & \cellcolor{teal!10}0.61 \\
& \multicolumn{1}{l}{RLVR} \\        
& \hspace{0.4em}w/ Confidence & \textbf{0.55} & 0.43 & 0.37 & \cellcolor{teal!10}0.36 \\
& \hspace{0.4em}w/ Probability & \textbf{0.55} & 0.57 & 0.43 & \cellcolor{teal!10}0.43 \\
& \hspace{0.4em}w/ Post-hoc & \textbf{0.55} & \textbf{0.79} & 0.25 & \cellcolor{teal!10}0.13 \\
& Behavioral Calibration & 0.53 & 0.50 & 0.28 & \cellcolor{teal!10}0.17 \\
& RLCR & \textbf{0.55} & 0.62 & 0.25 & \cellcolor{teal!10}0.13 \\
& \cellcolor[gray]{.9}\textsc{RL-ARC} (ours) & \cellcolor[gray]{.9}\textbf{0.55} & \cellcolor[gray]{.9}0.63 & \cellcolor[gray]{.9}\textbf{0.24} & \cellcolor{teal!10}\textbf{0.10} \\

\midrule

\multirow{7}{*}{AIME24}
& Base & 0.03 & \textbf{0.78} & 0.89 & \cellcolor{teal!10}0.93 \\
& \multicolumn{1}{l}{RLVR} \\        
& \hspace{0.4em}w/ Confidence & 0.10 & 0.35 & 0.89 & \cellcolor{teal!10}0.89 \\
& \hspace{0.4em}w/ Probability & 0.10 & 0.64 & 0.84 & \cellcolor{teal!10}0.86 \\
& \hspace{0.4em}w/ Post-hoc & 0.10 & 0.68 & 0.37 & \cellcolor{teal!10}0.51 \\
& Behavioral Calibration & 0.10 & 0.52 & 0.45 & \cellcolor{teal!10}0.60 \\
& RLCR & 0.07 & 0.54 & 0.39 & \cellcolor{teal!10}0.57 \\
& \cellcolor[gray]{.9}\textsc{RL-ARC} (ours) & \cellcolor[gray]{.9}\textbf{0.13} & \cellcolor[gray]{.9}0.64 & \cellcolor[gray]{.9}\textbf{0.32} & \cellcolor{teal!10}\textbf{0.45} \\

\bottomrule
\end{tabular}
}
\caption{Evaluation results for accuracy and calibration metrics on Math (ID) benchmarks. The best results are highlighted in \textbf{boldface}. Here, we use Qwen2.5-7B as base model.}
\label{tab:appendix: qwen2.5 overall-resutls id}
\end{table*}

\begin{table*}[ht]
\tiny
\centering
\resizebox{\textwidth}{!}{
\begin{tabular}{llcccc}
\toprule
Benchmark & Method
& Acc.\,($\uparrow$)
& AUROC\,($\uparrow$)
& Brier\,($\downarrow$)
& \cellcolor{teal!10}ECE\,($\downarrow$) \\
\midrule

\multirow{8}{*}{Big-Math}
& Base & 0.73 & 0.52 & 0.24 & \cellcolor{teal!10}0.23 \\
& RLVR & & & & \\
& \hspace{0.4em}w/ Confidence & \textbf{0.78} & 0.67 & 0.21 & \cellcolor{teal!10}0.21 \\
& \hspace{0.4em}w/ Probability & \textbf{0.78} & 0.71 & 0.22 & \cellcolor{teal!10}0.22 \\
& \hspace{0.4em}w/ Post-hoc & \textbf{0.78} & \textbf{0.79} & 0.16 & \cellcolor{teal!10}0.13 \\
& Behavioral Calibration & \textbf{0.78} & 0.67 & 0.15 & \cellcolor{teal!10}0.08 \\
& RLCR & 0.77 & 0.78 & 0.14 & \cellcolor{teal!10}\textbf{0.03} \\
& \cellcolor[gray]{.9}\textsc{RL-ARC} (ours)
& \cellcolor[gray]{.9}0.77
& \cellcolor[gray]{.9}\textbf{0.79}
& \cellcolor[gray]{.9}\textbf{0.13}
& \cellcolor{teal!10}\textbf{0.03} \\

\midrule

\multirow{8}{*}{GSM8K}
& Base & 0.90 & 0.56 & 0.09 & \cellcolor{teal!10}0.06 \\
& RLVR & & & & \\
& \hspace{0.4em}w/ Confidence & 0.92 & 0.62 & 0.08 & \cellcolor{teal!10}0.08 \\
& \hspace{0.4em}w/ Probability & 0.92 & 0.64 & 0.08 & \cellcolor{teal!10}0.08 \\
& \hspace{0.4em}w/ Post-hoc & 0.92 & \textbf{0.76} & 0.12 & \cellcolor{teal!10}0.22 \\
& Behavioral Calibration & 0.93 & 0.62 & \textbf{0.06} & \cellcolor{teal!10}0.04 \\
& RLCR & 0.92 & 0.68 & 0.07 & \cellcolor{teal!10}0.07 \\
& \cellcolor[gray]{.9}\textsc{RL-ARC} (ours)
& \cellcolor[gray]{.9}\textbf{0.94}
& \cellcolor[gray]{.9}0.74
& \cellcolor[gray]{.9}\textbf{0.06}
& \cellcolor{teal!10}\textbf{0.03} \\

\midrule

\multirow{8}{*}{MATH-500}
& Base & 0.64 & 0.53 & 0.35 & \cellcolor{teal!10}0.35 \\
& RLVR & & & & \\
& \hspace{0.4em}w/ Confidence & 0.63 & 0.63 & 0.36 & \cellcolor{teal!10}0.36 \\
& \hspace{0.4em}w/ Probability & 0.63 & 0.74 & 0.35 & \cellcolor{teal!10}0.35 \\
& \hspace{0.4em}w/ Post-hoc & 0.63 & \textbf{0.78} & \textbf{0.23} & \cellcolor{teal!10}0.23 \\
& Behavioral Calibration & \textbf{0.66} & 0.56 & \textbf{0.23} & \cellcolor{teal!10}\textbf{0.18} \\
& RLCR & \textbf{0.66} & 0.66 & 0.24 & \cellcolor{teal!10}0.19 \\
& \cellcolor[gray]{.9}\textsc{RL-ARC} (ours)
& \cellcolor[gray]{.9}0.65
& \cellcolor[gray]{.9}0.68
& \cellcolor[gray]{.9}\textbf{0.23}
& \cellcolor{teal!10}\textbf{0.18} \\

\bottomrule
\end{tabular}
}
\caption{Evaluation results for accuracy and calibration metrics on Big-Math, GSM8K, and MATH-500 (ID) benchmarks. The best results are highlighted in \textbf{boldface}. We use Qwen3-8B as the base model.}
\label{tab:appendix: qwen3 overall-resutls id}
\end{table*}

\begin{table*}[ht]
\tiny
\centering
\resizebox{\textwidth}{!}{
\begin{tabular}{llcccc}
\toprule
Benchmark & Method
& Acc.\,($\uparrow$)
& AUROC\,($\uparrow$)
& Brier\,($\downarrow$)
& \cellcolor{teal!10}ECE\,($\downarrow$) \\
\midrule

\multirow{8}{*}{AMC23}
& Base & 0.68 & 0.40 & 0.32 & \cellcolor{teal!10}0.32 \\
& RLVR & & & & \\
& \hspace{0.4em}w/ Confidence & \textbf{0.90} & 0.57 & 0.10 & \cellcolor{teal!10}\textbf{0.09} \\
& \hspace{0.4em}w/ Probability & \textbf{0.90} & 0.79 & 0.11 & \cellcolor{teal!10}0.10 \\
& \hspace{0.4em}w/ Post-hoc & \textbf{0.90} & 0.93 & 0.08 & \cellcolor{teal!10}0.11 \\
& Behavioral Calibration & 0.78 & 0.76 & 0.15 & \cellcolor{teal!10}0.10 \\
& RLCR & 0.88 & 0.87 & 0.09 & \cellcolor{teal!10}0.10 \\
& \cellcolor[gray]{.9}\textsc{RL-ARC} (ours)
& \cellcolor[gray]{.9}0.85
& \cellcolor[gray]{.9}\textbf{0.98}
& \cellcolor[gray]{.9}\textbf{0.06}
& \cellcolor{teal!10}0.18 \\

\midrule

\multirow{8}{*}{AIME24}
& Base & 0.20 & 0.35 & 0.79 & \cellcolor{teal!10}0.79 \\
& RLVR & & & & \\
& \hspace{0.4em}w/ Confidence & 0.43 & 0.78 & 0.49 & \cellcolor{teal!10}0.52 \\
& \hspace{0.4em}w/ Probability & 0.43 & \textbf{0.93} & 0.38 & \cellcolor{teal!10}0.44 \\
& \hspace{0.4em}w/ Post-hoc & 0.43 & 0.78 & 0.26 & \cellcolor{teal!10}0.24 \\
& Behavioral Calibration & \textbf{0.47} & 0.83 & 0.23 & \cellcolor{teal!10}0.26 \\
& RLCR & 0.40 & 0.74 & 0.25 & \cellcolor{teal!10}0.27 \\
& \cellcolor[gray]{.9}\textsc{RL-ARC} (ours)
& \cellcolor[gray]{.9}0.43
& \cellcolor[gray]{.9}0.91
& \cellcolor[gray]{.9}\textbf{0.16}
& \cellcolor{teal!10}\textbf{0.23} \\

\midrule

\multirow{8}{*}{AIME25}
& Base & 0.20 & 0.68 & 0.73 & \cellcolor{teal!10}0.76 \\
& RLVR & & & & \\
& \hspace{0.4em}w/ Confidence & \textbf{0.33} & 0.71 & 0.62 & \cellcolor{teal!10}0.64 \\
& \hspace{0.4em}w/ Probability & \textbf{0.33} & 0.78 & 0.49 & \cellcolor{teal!10}0.55 \\
& \hspace{0.4em}w/ Post-hoc & \textbf{0.33} & \textbf{0.98} & \textbf{0.13} & \cellcolor{teal!10}\textbf{0.27} \\
& Behavioral Calibration & \textbf{0.33} & 0.85 & 0.28 & \cellcolor{teal!10}0.36 \\
& RLCR & \textbf{0.33} & 0.85 & 0.24 & \cellcolor{teal!10}0.29 \\
& \cellcolor[gray]{.9}\textsc{RL-ARC} (ours)
& \cellcolor[gray]{.9}\textbf{0.33}
& \cellcolor[gray]{.9}0.93
& \cellcolor[gray]{.9}0.14
& \cellcolor{teal!10}\textbf{0.27} \\

\bottomrule
\end{tabular}
}
\caption{Evaluation results for accuracy and calibration metrics on AMC23, AIME24, and AIME25 (ID) benchmarks. The best results are highlighted in \textbf{boldface}. We use Qwen3-8B as the base model.}
\label{tab:appendix: qwen3 overall-resutls id-part2}
\end{table*}

\begin{table*}[ht]
\centering
\tiny
\resizebox{\textwidth}{!}{
\begin{tabular}{llcccc}
\toprule
Benchmark & Method
& Acc.\,($\uparrow$)
& AUROC\,($\uparrow$)
& Brier\,($\downarrow$)
& \cellcolor{teal!10}ECE\,($\downarrow$) \\
\midrule

\multirow{4}{*}{Big-Math}
& Base
& 0.29 & 0.61 & 0.65
& \cellcolor{teal!10}0.67 \\

& RLVR
& 0.61 & 0.53 & 0.35
& \cellcolor{teal!10}0.34 \\

& RLCR
& 0.58 & 0.76 & 0.22
& \cellcolor{teal!10}0.14 \\

& \cellcolor[gray]{.9}\textsc{RL-ARC} (ours)
& \cellcolor[gray]{.9}\textbf{0.63}
& \cellcolor[gray]{.9}\textbf{0.77}
& \cellcolor[gray]{.9}\textbf{0.20}
& \cellcolor{teal!10}\textbf{0.08} \\

\midrule

\multirow{4}{*}{GSM8K}
& Base
& 0.38 & 0.55 & 0.60
& \cellcolor{teal!10}0.60 \\

& RLVR
& \textbf{0.82} & 0.51 & 0.16
& \cellcolor{teal!10}0.13 \\

& RLCR
& 0.81 & \textbf{0.74} & \textbf{0.14}
& \cellcolor{teal!10}\textbf{0.06} \\

& \cellcolor[gray]{.9}\textsc{RL-ARC} (ours)
& \cellcolor[gray]{.9}0.81
& \cellcolor[gray]{.9}0.73
& \cellcolor[gray]{.9}\textbf{0.14}
& \cellcolor{teal!10}0.09 \\

\midrule

\multirow{4}{*}{MATH-500}
& Base
& 0.28 & 0.58 & 0.69
& \cellcolor{teal!10}0.70 \\

& RLVR
& \textbf{0.58} & 0.52 & 0.39
& \cellcolor{teal!10}0.38 \\

& RLCR
& 0.53 & 0.72 & 0.27
& \cellcolor{teal!10}0.23 \\

& \cellcolor[gray]{.9}\textsc{RL-ARC} (ours)
& \cellcolor[gray]{.9}\textbf{0.58}
& \cellcolor[gray]{.9}\textbf{0.78}
& \cellcolor[gray]{.9}\textbf{0.21}
& \cellcolor{teal!10}\textbf{0.17} \\

\midrule

\multirow{4}{*}{AMC23}
& Base
& 0.47 & 0.56 & 0.49
& \cellcolor{teal!10}0.50 \\

& RLVR
& \textbf{0.70} & 0.43 & 0.28
& \cellcolor{teal!10}\textbf{0.25} \\

& RLCR
& 0.45 & 0.85 & 0.36
& \cellcolor{teal!10}0.41 \\

& \cellcolor[gray]{.9}\textsc{RL-ARC} (ours)
& \cellcolor[gray]{.9}0.65
& \cellcolor[gray]{.9}\textbf{0.91}
& \cellcolor[gray]{.9}\textbf{0.27}
& \cellcolor{teal!10}0.29 \\

\midrule

\multirow{4}{*}{AIME24}
& Base
& 0.23 & 0.75 & 0.66
& \cellcolor{teal!10}0.69 \\

& RLVR
& \textbf{0.27} & 0.57 & 0.63
& \cellcolor{teal!10}0.65 \\

& RLCR
& 0.23 & \textbf{0.96} & \textbf{0.47}
& \cellcolor{teal!10}\textbf{0.58} \\

& \cellcolor[gray]{.9}\textsc{RL-ARC} (ours)
& \cellcolor[gray]{.9}0.17
& \cellcolor[gray]{.9}0.88
& \cellcolor[gray]{.9}0.51
& \cellcolor{teal!10}0.62 \\

\bottomrule
\end{tabular}
}

\caption{
Evaluation results for accuracy and calibration metrics on Math (ID)
benchmarks. The best results are highlighted in \textbf{boldface}.
Here, we use Llama-8B distilled from DeepSeek-R1 as the base model.
}
\label{tab:appendix:deepseekllama-overall-results-id}
\end{table*}

\begin{table*}[ht]
\centering
\tiny
\resizebox{\textwidth}{!}{
\begin{tabular}{llcccc}
\toprule
Benchmark & Method
& Acc.\,($\uparrow$) 
& AUROC\,($\uparrow$) 
& Brier\,($\downarrow$) 
& \cellcolor{teal!10}ECE\,($\downarrow$) \\
\midrule

\multirow{7}{*}{StrategyQA}
& Base & 0.68 & \textbf{0.58} & 0.27 & \cellcolor{teal!10}0.23 \\
& \multicolumn{1}{l}{RLVR} \\        
& \hspace{0.4em}w/ Confidence & 0.70 & 0.48 & 0.29 & \cellcolor{teal!10}0.29 \\
& \hspace{0.4em}w/ Probability & 0.70 & 0.56 & 0.30 & \cellcolor{teal!10}0.29 \\
& \hspace{0.4em}w/ Post-hoc & 0.70 & 0.45 & 0.25 & \cellcolor{teal!10}0.19 \\
& Behavioral Calibration & 0.72 & 0.53 & \textbf{0.20} & \cellcolor{teal!10}\textbf{0.04} \\
& RLCR & 0.68 & 0.49 & 0.24 & \cellcolor{teal!10}0.09 \\
& \cellcolor[gray]{.9}\textsc{RL-ARC} (ours) & \cellcolor[gray]{.9}\textbf{0.74} & \cellcolor[gray]{.9}0.49 & \cellcolor[gray]{.9}0.23 & \cellcolor{teal!10}0.19 \\

\midrule

\multirow{7}{*}{HotpotQA}
& Base & 0.39 & 0.54 & 0.53 & \cellcolor{teal!10}0.54 \\
& \multicolumn{1}{l}{RLVR} \\        
& \hspace{0.4em}w/ Confidence & 0.47 & 0.50 & 0.53 & \cellcolor{teal!10}0.53 \\
& \hspace{0.4em}w/ Probability & 0.47 & \textbf{0.62} & 0.49 & \cellcolor{teal!10}0.50 \\
& \hspace{0.4em}w/ Post-hoc & 0.47 & 0.48 & 0.30 & \cellcolor{teal!10}0.18 \\
& Behavioral Calibration & \textbf{0.49} & 0.52 & 0.27 & \cellcolor{teal!10}0.14 \\
& RLCR & 0.47 & 0.56 & 0.26 & \cellcolor{teal!10}0.15 \\
& \cellcolor[gray]{.9}\textsc{RL-ARC} (ours) & \cellcolor[gray]{.9}0.48 & \cellcolor[gray]{.9}0.60 & \cellcolor[gray]{.9}\textbf{0.24} & \cellcolor{teal!10}\textbf{0.06} \\

\midrule

\multirow{7}{*}{GPQA}
& Base & 0.37 & \textbf{0.54} & 0.51 & \cellcolor{teal!10}0.52 \\
& \multicolumn{1}{l}{RLVR} \\        
& \hspace{0.4em}w/ Confidence & \textbf{0.48} & 0.52 & 0.52 & \cellcolor{teal!10}0.52 \\
& \hspace{0.4em}w/ Probability & \textbf{0.48} & 0.45 & 0.51 & \cellcolor{teal!10}0.48 \\
& \hspace{0.4em}w/ Post-hoc & \textbf{0.48} & 0.52 & \textbf{0.27} & \cellcolor{teal!10}\textbf{0.12} \\
& Behavioral Calibration & 0.42 & 0.50 & 0.31 & \cellcolor{teal!10}0.25 \\
& RLCR & 0.40 & 0.53 & 0.30 & \cellcolor{teal!10}0.24 \\
& \cellcolor[gray]{.9}\textsc{RL-ARC} (ours) & \cellcolor[gray]{.9}0.38 & \cellcolor[gray]{.9}0.53 & \cellcolor[gray]{.9}\textbf{0.27} & \cellcolor{teal!10}0.17 \\

\bottomrule
\end{tabular}
}
\caption{Evaluation results for accuracy and calibration metrics on complex reasoning (OOD) benchmarks. The best results are highlighted in \textbf{boldface}. Here, we use Qwen2.5-7B as base model.}
\label{tab:appendix: qwen2.5 overall-resutls ood complex reasoning}
\end{table*}

\begin{table*}[ht]
\centering
\tiny
\resizebox{\textwidth}{!}{
\begin{tabular}{llcccc}
\toprule
Benchmark & Method
& Acc.\,($\uparrow$) 
& AUROC\,($\uparrow$) 
& Brier\,($\downarrow$) 
& \cellcolor{teal!10}ECE\,($\downarrow$) \\
\midrule

\multirow{7}{*}{SimpleQA}
& Base & 0.13 & 0.50 & 0.78 & \cellcolor{teal!10}0.81 \\
& \multicolumn{1}{l}{RLVR} \\        
& \hspace{0.4em}w/ Confidence & \textbf{0.14} & 0.51 & 0.81 & \cellcolor{teal!10}0.83 \\
& \hspace{0.4em}w/ Probability &\textbf{0.14}  &0.46 &0.79 & \cellcolor{teal!10}0.81 \\
& \hspace{0.4em}w/ Post-hoc & \textbf{0.14} &0.47 &0.50 & \cellcolor{teal!10}0.60 \\
& Behavioral Calibration & 0.12 & 0.50 & 0.44 & \cellcolor{teal!10}0.58 \\
& RLCR & 0.12  & \textbf{0.52} & 0.37 & \cellcolor{teal!10}0.51  \\
& \cellcolor[gray]{.9}\textsc{RL-ARC} (ours) & \cellcolor[gray]{.9}0.13 & \cellcolor[gray]{.9}0.45 & \cellcolor[gray]{.9}\textbf{0.32} & \cellcolor{teal!10}\textbf{0.43} \\

\midrule

\multirow{7}{*}{NQ-Open}
& Base &0.47  &0.51 &0.48 & \cellcolor{teal!10}0.48 \\
& \multicolumn{1}{l}{RLVR} \\        
& \hspace{0.4em}w/ Confidence &0.47  &0.50 &0.50 & \cellcolor{teal!10}0.50 \\
& \hspace{0.4em}w/ Probability &0.47  &0.48 &0.45 & \cellcolor{teal!10}0.40 \\
& \hspace{0.4em}w/ Post-hoc &0.47  &\textbf{0.56} &0.31 & \cellcolor{teal!10}0.25 \\
& Behavioral Calibration & \textbf{0.49 }& 0.50 & 0.29 & \cellcolor{teal!10}0.21 \\
& RLCR &\textbf{0.49}  &0.54 &0.27 & \cellcolor{teal!10}0.14 \\
& \cellcolor[gray]{.9}\textsc{RL-ARC} (ours) & \cellcolor[gray]{.9}\textbf{0.49} & \cellcolor[gray]{.9}\textbf{0.56} & \cellcolor[gray]{.9}\textbf{0.26} & \cellcolor{teal!10}\textbf{0.09} \\

\midrule

\multirow{7}{*}{TriviaQA}
& Base & 0.57 &0.52 &0.39 & \cellcolor{teal!10}0.38 \\
& \multicolumn{1}{l}{RLVR} \\        
& \hspace{0.4em}w/ Confidence &0.60  &0.48 &0.38 & \cellcolor{teal!10}0.36 \\
& \hspace{0.4em}w/ Probability &0.60  &0.50 &0.37 & \cellcolor{teal!10}0.36 \\
& \hspace{0.4em}w/ Post-hoc &0.60  &0.58 &0.25 & \cellcolor{teal!10}0.10 \\
& Behavioral Calibration & 0.58 & 0.50 & 0.26 & \cellcolor{teal!10}0.12 \\
& RLCR &0.60  &0.56 &0.24 & \cellcolor{teal!10}0.05 \\
& \cellcolor[gray]{.9}\textsc{RL-ARC} (ours) & \cellcolor[gray]{.9}\textbf{0.61} & \cellcolor[gray]{.9}\textbf{0.60} & \cellcolor[gray]{.9}\textbf{0.23} & \cellcolor{teal!10}\textbf{0.04} \\

\bottomrule
\end{tabular}
}
\caption{Evaluation results for accuracy and calibration metrics on factual question answering (OOD) benchmarks. The best results are highlighted in \textbf{boldface}. Here, we use Qwen2.5-7B as base model.}
\label{tab:appendix: qwen2.5 overall-resutls ood factual}
\end{table*}

\clearpage

\begin{table*}[ht]
\centering
\tiny
\resizebox{\textwidth}{!}{
\begin{tabular}{llcccc}
\toprule
Benchmark & Method
& Acc.\,($\uparrow$) 
& AUROC\,($\uparrow$) 
& Brier\,($\downarrow$) 
& \cellcolor{teal!10}ECE\,($\downarrow$) \\
\midrule

\multirow{4}{*}{GPQA}
& Base
& 0.37 & \textbf{0.54} & 0.51
& \cellcolor{teal!10}0.52 \\

& RLVR
& 0.39 & 0.50 & 0.61
& \cellcolor{teal!10}0.61 \\

& RLCR
& \textbf{0.40} & \textbf{0.54} & 0.29
& \cellcolor{teal!10}0.22 \\

& \cellcolor[gray]{.9}\textsc{RL-ARC} (ours)
& \cellcolor[gray]{.9}\textbf{0.40}
& \cellcolor[gray]{.9}0.52
& \cellcolor[gray]{.9}\textbf{0.26}
& \cellcolor{teal!10}\textbf{0.15} \\

\midrule

\multirow{4}{*}{GSM8K}
& Base
& \textbf{0.73} & 0.52 & 0.25
& \cellcolor{teal!10}0.22 \\

& RLVR
& 0.49 & 0.50 & 0.51
& \cellcolor{teal!10}0.51 \\

& RLCR
& 0.66 & 0.57 & 0.22
& \cellcolor{teal!10}\textbf{0.08} \\

& \cellcolor[gray]{.9}\textsc{RL-ARC} (ours)
& \cellcolor[gray]{.9}0.72
& \cellcolor[gray]{.9}\textbf{0.65}
& \cellcolor[gray]{.9}\textbf{0.21}
& \cellcolor{teal!10}0.15 \\

\midrule

\multirow{4}{*}{MATH-500}
& Base
& \textbf{0.45} & 0.58 & 0.50
& \cellcolor{teal!10}0.51 \\

& RLVR
& 0.30 & 0.50 & 0.70
& \cellcolor{teal!10}0.70 \\

& RLCR
& 0.33 & 0.55 & 0.30
& \cellcolor{teal!10}0.28 \\

& \cellcolor[gray]{.9}\textsc{RL-ARC} (ours)
& \cellcolor[gray]{.9}0.39
& \cellcolor[gray]{.9}\textbf{0.60}
& \cellcolor[gray]{.9}\textbf{0.27}
& \cellcolor{teal!10}\textbf{0.18} \\

\bottomrule
\end{tabular}
}

\caption{
Evaluation results for accuracy and calibration metrics on complex reasoning (OOD)
benchmarks. The best results are highlighted in \textbf{boldface}.
Here, we use Qwen2.5-7B as the base model.
}
\label{tab:appendix:qwenhotpot-overall-results-ood-complex}
\end{table*}

\begin{table*}[ht]
\centering
\tiny
\resizebox{\textwidth}{!}{
\begin{tabular}{llcccc}
\toprule
Benchmark & Method
& Acc.\,($\uparrow$) 
& AUROC\,($\uparrow$) 
& Brier\,($\downarrow$) 
& \cellcolor{teal!10}ECE\,($\downarrow$) \\
\midrule

\multirow{4}{*}{SimpleQA}
& Base
& \textbf{0.13} & 0.50 & 0.78
& \cellcolor{teal!10}0.81 \\

& RLVR
& 0.11 & 0.50 & 0.89
& \cellcolor{teal!10}0.89 \\

& RLCR
& 0.12 & \textbf{0.54} & 0.33
& \cellcolor{teal!10}0.45 \\

& \cellcolor[gray]{.9}\textsc{RL-ARC} (ours)
& \cellcolor[gray]{.9}0.12
& \cellcolor[gray]{.9}0.51
& \cellcolor[gray]{.9}\textbf{0.24}
& \cellcolor{teal!10}\textbf{0.35} \\

\midrule

\multirow{4}{*}{TriviaQA}
& Base
& 0.57 & 0.52 & 0.39
& \cellcolor{teal!10}0.38 \\

& RLVR
& \textbf{0.60} & 0.50 & 0.40
& \cellcolor{teal!10}0.40 \\

& RLCR
& 0.59 & \textbf{0.63} & \textbf{0.23}
& \cellcolor{teal!10}\textbf{0.05} \\

& \cellcolor[gray]{.9}\textsc{RL-ARC} (ours)
& \cellcolor[gray]{.9}0.59
& \cellcolor[gray]{.9}0.59
& \cellcolor[gray]{.9}0.24
& \cellcolor{teal!10}0.07 \\

\midrule

\multirow{4}{*}{NQ-Open}
& Base
& 0.47 & 0.51 & 0.48
& \cellcolor{teal!10}0.48 \\

& RLVR
& 0.49 & 0.50 & 0.51
& \cellcolor{teal!10}0.51 \\

& RLCR
& \textbf{0.51} & \textbf{0.57} & 0.26
& \cellcolor{teal!10}0.13 \\

& \cellcolor[gray]{.9}\textsc{RL-ARC} (ours)
& \cellcolor[gray]{.9}0.50
& \cellcolor[gray]{.9}0.55
& \cellcolor[gray]{.9}\textbf{0.25}
& \cellcolor{teal!10}\textbf{0.04} \\

\bottomrule
\end{tabular}
}

\caption{
Evaluation results for accuracy and calibration metrics on factual knowledge (OOD)
benchmarks. The best results are highlighted in \textbf{boldface}.
Here, we use Qwen2.5-7B as the base model.
}
\label{tab:appendix:qwen_hotpotqa_overall-results-ood-factual}
\end{table*}

\begin{table*}[ht]
\centering
\tiny
\resizebox{\textwidth}{!}{
\begin{tabular}{llcccc}
\toprule
Benchmark & Method
& Acc.\,($\uparrow$) 
& AUROC\,($\uparrow$) 
& Brier\,($\downarrow$) 
& \cellcolor{teal!10}ECE\,($\downarrow$) \\
\midrule

\multirow{7}{*}{StrategyQA}
& Base &0.71 &0.62  &0.24  & \cellcolor{teal!10}0.20 \\
& \multicolumn{1}{l}{RLVR} \\        
& \hspace{0.4em}w/ Confidence &0.70  & 0.56 &0.27  & \cellcolor{teal!10}0.26 \\
& \hspace{0.4em}w/ Probability &0.70 &0.49  &0.30  & \cellcolor{teal!10} 0.30\\
& \hspace{0.4em}w/ Post-hoc &0.70 &0.55  &0.30  & \cellcolor{teal!10}0.31 \\
& Behavioral Calibration & 0.69 & 0.60 & 0.24 & \cellcolor{teal!10}0.17 \\
& RLCR &\textbf{0.72} & 0.70 &0.19  & \cellcolor{teal!10}0.04 \\
& \cellcolor[gray]{.9}\textsc{RL-ARC} (ours) & \cellcolor[gray]{.9}0.71 & \cellcolor[gray]{.9}\textbf{0.71} & \cellcolor[gray]{.9}\textbf{0.18} & \cellcolor{teal!10}\textbf{0.03} \\

\midrule

\multirow{7}{*}{HotpotQA}
& Base & 0.61& 0.56 &0.34  & \cellcolor{teal!10}0.33 \\
& \multicolumn{1}{l}{RLVR} \\        
& \hspace{0.4em}w/ Confidence &\textbf{0.62} &0.54  &0.38  & \cellcolor{teal!10}0.38 \\
& \hspace{0.4em}w/ Probability &\textbf{0.62} &\textbf{0.59}  &0.38  & \cellcolor{teal!10}0.38 \\
& \hspace{0.4em}w/ Post-hoc &\textbf{0.62} &0.55  &\textbf{0.27}  & \cellcolor{teal!10}\textbf{0.18} \\
& Behavioral Calibration & 0.61 & 0.53 & 0.31 & \cellcolor{teal!10}0.28 \\
& RLCR &\textbf{0.62} & 0.56 &0.30  & \cellcolor{teal!10}0.26 \\
& \cellcolor[gray]{.9}\textsc{RL-ARC} (ours) & \cellcolor[gray]{.9}0.61 & \cellcolor[gray]{.9}0.57 & \cellcolor[gray]{.9}0.29 & \cellcolor{teal!10}0.24 \\

\midrule

\multirow{7}{*}{GPQA}
& Base &0.54 &0.63  &0.25  & \cellcolor{teal!10}0.20 \\
& \multicolumn{1}{l}{RLVR} \\        
& \hspace{0.4em}w/ Confidence &0.53 & 0.52 &0.42  & \cellcolor{teal!10}0.42 \\
& \hspace{0.4em}w/ Probability &0.53 &0.48  & 0.46 & \cellcolor{teal!10}0.46 \\
& \hspace{0.4em}w/ Post-hoc &0.53 & 0.60 &0.29  & \cellcolor{teal!10}0.21 \\
& Behavioral Calibration & \textbf{0.58} & 0.61 & 0.28 & \cellcolor{teal!10}0.20 \\
& RLCR & 0.51&0.63  &0.25  & \cellcolor{teal!10}0.12 \\
& \cellcolor[gray]{.9}\textsc{RL-ARC} (ours) & \cellcolor[gray]{.9}0.56 & \cellcolor[gray]{.9}\textbf{0.64} & \cellcolor[gray]{.9}\textbf{0.23} & \cellcolor{teal!10}\textbf{0.05} \\

\bottomrule
\end{tabular}
}
\caption{Evaluation results for accuracy and calibration metrics on complex reasoning (OOD) benchmarks. The best results are highlighted in \textbf{boldface}. Here, we use Qwen3-8B as base model.}
\label{tab:appendix: qwen3 overall-resutls ood complex resonign}
\end{table*}

\begin{table*}[ht]
\centering
\tiny
\resizebox{\textwidth}{!}{
\begin{tabular}{llcccc}
\toprule
Benchmark & Method
& Acc.\,($\uparrow$) 
& AUROC\,($\uparrow$) 
& Brier\,($\downarrow$) 
& \cellcolor{teal!10}ECE\,($\downarrow$) \\
\midrule

\multirow{7}{*}{SimpleQA}
& Base &\textbf{0.12} &0.63  &0.51  & \cellcolor{teal!10}0.62 \\
& \multicolumn{1}{l}{RLVR} \\        
& \hspace{0.4em}w/ Confidence &0.10  &0.53  &0.78  & \cellcolor{teal!10}0.83 \\
& \hspace{0.4em}w/ Probability &0.10 &0.38  &0.80  & \cellcolor{teal!10}0.81 \\
& \hspace{0.4em}w/ Post-hoc & 0.10 &0.56  & 0.29 & \cellcolor{teal!10}0.38 \\
& Behavioral Calibration & 0.11 & 0.60 & 0.50 & \cellcolor{teal!10}0.61 \\
& RLCR &0.11 &0.60  &0.33  & \cellcolor{teal!10}0.46 \\
& \cellcolor[gray]{.9}\textsc{RL-ARC} (ours) & \cellcolor[gray]{.9}0.11 & \cellcolor[gray]{.9}\textbf{0.65} & \cellcolor[gray]{.9}\textbf{0.24} & \cellcolor{teal!10}\textbf{0.35} \\

\midrule

\multirow{7}{*}{NQ-Open}
& Base &0.49 & 0.71 &0.39  & \cellcolor{teal!10}0.40 \\
& \multicolumn{1}{l}{RLVR} \\        
& \hspace{0.4em}w/ Confidence &0.33 &0.57  &0.60  & \cellcolor{teal!10}0.61 \\
& \hspace{0.4em}w/ Probability &0.33 &0.49  & 0.41 & \cellcolor{teal!10}0.40 \\
& \hspace{0.4em}w/ Post-hoc &0.33 & 0.52 &0.30  & \cellcolor{teal!10}\textbf{0.21} \\
& Behavioral Calibration & \textbf{0.50} & 0.64 & 0.35 & \cellcolor{teal!10}0.34 \\
& RLCR &0.49 &0.70  &0.30  & \cellcolor{teal!10}0.27 \\
& \cellcolor[gray]{.9}\textsc{RL-ARC} (ours) & \cellcolor[gray]{.9}\textbf{0.50} & \cellcolor[gray]{.9}\textbf{0.73} & \cellcolor[gray]{.9}\textbf{0.25} & \cellcolor{teal!10}\textbf{0.21} \\

\midrule

\multirow{7}{*}{TriviaQA}
& Base &\textbf{0.61} &0.76  &0.29  & \cellcolor{teal!10}0.29 \\
& \multicolumn{1}{l}{RLVR} \\        
& \hspace{0.4em}w/ Confidence &0.55 &0.61  &0.40  & \cellcolor{teal!10}0.40 \\
& \hspace{0.4em}w/ Probability &0.55 &0.58  &0.34  & \cellcolor{teal!10}0.34 \\
& \hspace{0.4em}w/ Post-hoc &0.55 &0.63  &0.25  & \cellcolor{teal!10}\textbf{0.14} \\
& Behavioral Calibration & 0.59 & 0.64 & 0.29 & \cellcolor{teal!10}0.27 \\
& RLCR &\textbf{0.61} &0.78  &0.22  & \cellcolor{teal!10}0.18 \\
& \cellcolor[gray]{.9}\textsc{RL-ARC} (ours) & \cellcolor[gray]{.9}0.60 & \cellcolor[gray]{.9}\textbf{0.81} & \cellcolor[gray]{.9}\textbf{0.20} & \cellcolor{teal!10}\textbf{0.14} \\
\bottomrule
\end{tabular}
}
\caption{Evaluation results for accuracy and calibration metrics on factual question answering (OOD) benchmarks. The best results are highlighted in \textbf{boldface}. Here, we use Qwen3-8B as base model.}
\label{tab:appendix: qwen3 overall-resutls ood factual}
\end{table*}

\clearpage

\begin{table*}[ht]
\centering
\tiny
\resizebox{\textwidth}{!}{
\begin{tabular}{llcccc}
\toprule
Benchmark & Method
& Acc.\,($\uparrow$) 
& AUROC\,($\uparrow$) 
& Brier\,($\downarrow$) 
& \cellcolor{teal!10}ECE\,($\downarrow$) \\
\midrule

\multirow{4}{*}{StrategyQA}
& Base
& 0.64 & \textbf{0.61} & 0.27
& \cellcolor{teal!10}0.23 \\

& RLVR
& 0.64 & 0.50 & 0.32
& \cellcolor{teal!10}0.29 \\

& RLCR
& 0.64 & \textbf{0.61} & \textbf{0.24}
& \cellcolor{teal!10}0.14 \\

& \cellcolor[gray]{.9}\textsc{RL-ARC} (ours)
& \cellcolor[gray]{.9}\textbf{0.65}
& \cellcolor[gray]{.9}0.59
& \cellcolor[gray]{.9}\textbf{0.24}
& \cellcolor{teal!10}\textbf{0.12} \\

\midrule

\multirow{4}{*}{HotpotQA}
& Base
& \textbf{0.47} & 0.55 & 0.49
& \cellcolor{teal!10}0.50 \\

& RLVR
& 0.27 & 0.50 & 0.66
& \cellcolor{teal!10}0.68 \\

& RLCR
& 0.38 & \textbf{0.63} & 0.38
& \cellcolor{teal!10}0.39 \\

& \cellcolor[gray]{.9}\textsc{RL-ARC} (ours)
& \cellcolor[gray]{.9}0.40
& \cellcolor[gray]{.9}\textbf{0.63}
& \cellcolor[gray]{.9}\textbf{0.30}
& \cellcolor{teal!10}\textbf{0.25} \\

\midrule

\multirow{4}{*}{GPQA}
& Base
& \textbf{0.67} & 0.55 & 0.28
& \cellcolor{teal!10}0.25 \\

& RLVR
& 0.56 & 0.55 & 0.38
& \cellcolor{teal!10}0.37 \\

& RLCR
& 0.57 & \textbf{0.62} & \textbf{0.24}
& \cellcolor{teal!10}\textbf{0.09} \\

& \cellcolor[gray]{.9}\textsc{RL-ARC} (ours)
& \cellcolor[gray]{.9}0.58
& \cellcolor[gray]{.9}0.60
& \cellcolor[gray]{.9}0.25
& \cellcolor{teal!10}\textbf{0.09} \\

\bottomrule
\end{tabular}
}

\caption{
Evaluation results for accuracy and calibration metrics on complex reasoning (OOD)
benchmarks. The best results are highlighted in \textbf{boldface}.
Here, we use Llama-8B distilled from DeepSeek-R1 as the base model.
}
\label{tab:appendix:deepseekllama-overall-results-ood-complex}
\end{table*}

\begin{table*}[ht]
\centering
\tiny
\resizebox{\textwidth}{!}{
\begin{tabular}{llcccc}
\toprule
Benchmark & Method
& Acc.\,($\uparrow$) 
& AUROC\,($\uparrow$) 
& Brier\,($\downarrow$) 
& \cellcolor{teal!10}ECE\,($\downarrow$) \\
\midrule

\multirow{4}{*}{SimpleQA}
& Base
& 0.10 & 0.57 & 0.65
& \cellcolor{teal!10}0.75 \\

& RLVR
& 0.10 & 0.51 & 0.80
& \cellcolor{teal!10}0.84 \\

& RLCR
& \textbf{0.12} & 0.62 & 0.37
& \cellcolor{teal!10}0.52 \\

& \cellcolor[gray]{.9}\textsc{RL-ARC} (ours)
& \cellcolor[gray]{.9}0.10
& \cellcolor[gray]{.9}\textbf{0.65}
& \cellcolor[gray]{.9}\textbf{0.17}
& \cellcolor{teal!10}\textbf{0.28} \\

\midrule

\multirow{4}{*}{NQ-Open}
& Base
& 0.40 & \textbf{0.68} & 0.46
& \cellcolor{teal!10}0.49 \\

& RLVR
& \textbf{0.41} & 0.51 & 0.53
& \cellcolor{teal!10}0.54 \\

& RLCR
& 0.38 & 0.67 & 0.32
& \cellcolor{teal!10}0.31 \\

& \cellcolor[gray]{.9}\textsc{RL-ARC} (ours)
& \cellcolor[gray]{.9}0.40
& \cellcolor[gray]{.9}0.67
& \cellcolor[gray]{.9}\textbf{0.22}
& \cellcolor{teal!10}\textbf{0.12} \\

\midrule

\multirow{4}{*}{TriviaQA}
& Base
& \textbf{0.48} & \textbf{0.71} & 0.41
& \cellcolor{teal!10}0.43 \\

& RLVR
& 0.46 & 0.51 & 0.48
& \cellcolor{teal!10}0.48 \\

& RLCR
& 0.45 & 0.70 & \textbf{0.28}
& \cellcolor{teal!10}\textbf{0.23} \\

& \cellcolor[gray]{.9}\textsc{RL-ARC} (ours)
& \cellcolor[gray]{.9}0.47
& \cellcolor[gray]{.9}\textbf{0.71}
& \cellcolor[gray]{.9}0.31
& \cellcolor{teal!10}0.29 \\

\bottomrule
\end{tabular}
}

\caption{
Evaluation results for accuracy and calibration metrics on factual knowledge (OOD)
benchmarks. The best results are highlighted in \textbf{boldface}.
Here, we use Llama-8B distilled from DeepSeek-R1 as the base model.
}
\label{tab:appendix:deepseekllama-overall-results-ood-factual}
\end{table*}

\end{document}